\documentclass[lettersize,journal]{IEEEtran}
\usepackage{multirow}
\usepackage{amsfonts,amsmath,amssymb,mathtools}
\usepackage{graphicx}
\usepackage{epstopdf}
\usepackage{xurl}
\usepackage{cite}
\usepackage[caption=false,font=footnotesize,subrefformat=parens,labelformat=parens]{subfig}
\usepackage{comment}
\usepackage[ruled,vlined]{algorithm2e}
\usepackage{algpseudocode}
\usepackage{tikz}
\usetikzlibrary{arrows,intersections,decorations,plotmarks,math,patterns}
\usepackage{pgfplots}
\pgfplotsset{compat=1.17}
\usepackage{color}
\usepackage[colorinlistoftodos]{todonotes}

\usepackage{hyperref}

\definecolor{COL11}{RGB}{215,25,28}
\definecolor{COL12}{RGB}{253,174,97}
\definecolor{COL13}{RGB}{146,197,222}
\def\BibTeX{{\rm B\kern-.05em{\sc i\kern-.025em b}\kern-.08em
    T\kern-.1667em\lower.7ex\hbox{E}\kern-.125emX}}
\usepackage{balance}
\usepackage{booktabs}

\begin{document}

\title{Explainable Artificial Intelligence based Soft Evaluation Indicator for Arc Fault Diagnosis} 

\author{
Qianchao Wang$^{1}$ ,  Yuxuan Ding$^{1\dag}$, Chuanzhen Jia$^{1\dag}$, Zhe Li $^{2}$, Yaping Du $^{1}$ 

\thanks{Yuxuan Ding and Chuanzhen Jia are the corresponding authors}
\thanks{$^{1}$ Qianchao Wang, Chuanzhen Jia, Yuxuan Ding, and Yaping Du are with the Department of Building Environment and Energy Engineering, Hong Kong Polytechnic University, Hung Hom, Hong Kong. (qianchao.wang@polyu.edu.hk; chuanzhen.jia@connect.polyu.hk; yx.ding@connect.polyu.hk; Ya-ping.du@polyu.edu.hk;  )} 
\thanks{$^{2}$ Zhe Li is with Shenzhen Power Supply Bureau Co., Ltd, Guangzhou, China. (19109887r@connect.polyu.hk)}

\thanks{This work was supported by the Science and Technology Project of China Southern Power Grid Co., Ltd. ``Research on Data Set Platform and Testing Standard for Low Voltage Electricity Safety Hazar''(090000KC24110001).} 

}

\maketitle

\begin{abstract}

Novel AI-based arc fault diagnosis models have demonstrated outstanding performance in terms of classification accuracy. However, an inherent problem is whether these models can actually be trusted to find arc faults. In this light, this work proposes a soft evaluation indicator that explains the outputs of arc fault diagnosis models, by defining the
the correct explanation of arc faults and leveraging Explainable Artificial Intelligence and real arc fault experiments. Meanwhile, a lightweight balanced neural network is proposed to guarantee competitive accuracy and soft feature extraction score.
In our experiments, several traditional machine learning methods and deep learning methods across two arc fault datasets with different sample times and noise levels are utilized to test the effectiveness of the soft evaluation indicator. Through this approach, the arc fault diagnosis models are easy to understand and trust, allowing practitioners to make informed and
trustworthy decisions.


\end{abstract}


\section{Introduction}\label{Introduction}
\IEEEPARstart{W}{ith} the deepening of the electrification of buildings and transportation, arc faults have become an essential problem in power systems, since they can ignite surrounding materials, leading to fires that often go undetected~\cite{jiang2022machine} and posing serious threats to people and property~\cite{thakur2023advancements}. Meanwhile, the arc faults will reduce the current of the circuit, which causes the conventional over-current and leakage current protection devices to fail to detect the fault~\cite{jiang2021coupling}. Therefore, many recent studies have designed many arc fault detection or classification methods to warn of the occurrence of arc faults in advance and avoid the tragedy of fire.

Apart from the physical-signal methods, the popular arc fault detection methods can be divided into two categories: time- and frequency-domain analysis and deep learning. The first category is usually combined with traditional machine learning (ML)~\cite{8970332}. By analyzing the input electrical signals from the perspective of the time domain and the frequency domain using the Fourier transform~\cite{8624391} or the wavelet transform~\cite{9950518,9301196}, an abundant feature pool can be well designed. According to Mayr’s model~\cite{9605243}, some simple but trustworthy features can be recognized as members of the feature pool, such as signal mean, variance, and range, despite the fact that they may be affected by load types. Moreover, some complex statistical biases are used as descriptors of time-domain signals as well, such as kurtosis~\cite{9756255}, skewness, entropy, norm~\cite{7287757}, zero current period, and maximum slip difference~\cite{8667831}. These features are manually selected by researchers on the basis of their own experiences, lacking a general specification to limit the application. When it comes to the frequency domain, the magnitudes of components are normally used as key features to decide the detection of arc faults. The performance of these features depends on the sampling frequency and the potential noise in the raw electrical signals. All these selected features are then input into machine learning classifiers, for instance, k-nearest neighbor classification~\cite{10130787}, ensemble machine learning methods~\cite{8970332}, and support vector machine~\cite{10141554}. The ML algorithms fully exploit the potential of the feature pool  to achieve Arc fault detection or classification. Therefore, the performance of arc fault detection or classification depends mainly on the rationality and accuracy of feature selection instead of the ML algorithms.

In another aspect, to avoid the complex feature extraction process for arc fault diagnosis, end-to-end classifications are also widely used, for example, deep learning methods~\cite{Park2021DeepLS,8571267}. The raw or pre-processed electrical signals (current or voltage) are directly leveraged as the input of deep neural networks, using the powerful feature extraction ability of networks to adaptively extract useful features. Some interesting networks are utilized in recent research, such as AutoEncoder (AE) ~\cite{9439848}, series arc fault neural network~\cite{10018466}, ArcNet~\cite{9392282}, and temporal convolution network (ArcNN) ~\cite{10054597}. The high detection and classification accuracy implies that these models can discover hidden arc faults through automatically extracted features, including using different sample times. This result-oriented thinking runs through the entire deep learning process, including the construction of additional features, the construction of model structure, and the selection of hyperparameters.

However, a core obstacle is that \textit{Do these methods really focus on the real arc fault?} From a naive causal inference perspective~\cite{hernan2010causal}, the precise and elegant feature extraction in the early stage of models contributes to the high performance of arc fault diagnosis models instead of a redundant and rough feature pool or blind pursuit of detection accuracy. Although more features mean a potentially higher detection accuracy, aimlessly expanding the feature pool and blindly pursuing accuracy will make practitioners lose confidence in the model. Therefore, while the arc fault diagnosis accuracy remains unchanged, we need to correctly evaluate whether the model has found the real fault, and based on this, consider whether to increase the model size or change the model architecture, especially when computing resources are limited.
A possible solution for this problem is to use explainable artificial intelligence (XAI)~\cite{ali2023explainable,van2024harnessing} to evaluate the aforementioned ML methods and form a unified evaluation system for AI-based arc fault diagnosis. XAI is normally used to increase the model's interpretability and transparency, based on game theory, gradient-based methods, or attribution techniques. It has been widely used in many aspects, such as building energy management~\cite{chen2023interpretable}, load disaggregation~\cite{10198359}, and power quality disturbances classification~\cite{10287853}. The core idea of XAI is consistent with causal inference and engineering requirements, helping the domain experts and researchers better understand and evaluate the inner operation of models.

In this light, the primary objective of this work is to fully utilize the ability of XAI to establish a novel and elegant posterior evaluation system for AI-based arc fault diagnosis to improve practitioners’ confidence in models. We proposed a unified soft evaluation indicator, named explainable soft evaluation indicator (XSEI), for both feature pool-based and automatic feature extraction-based methods by defining the correct explanation of arc faults and leveraging SHapley Additive exPlanations (SHAP)~\cite{lundberg2017unified,chen2023algorithms} and occlusion sensitivity~\cite{zeiler2014visualizing} during a new validation process. The calculated evaluation indicators can be used as the basis for model selection or considered as the weights of ensemble machine learning, improving the transparency of the output. Notably, this method holds the potential for adaptation to other domains within industrial diagnostics, and the fault detection accuracy is not our primary pursuit. To demonstrate the universality of evaluation indicators, we conduct an arc fault experiment and evaluate several popular AI-based arc fault diagnosis methods, including traditional machine learning methods and deep learning methods. In summary, the key contributions of this paper encompass:
\begin{enumerate} 
    \item A definition for the correct explanation of machine learning-based arc fault diagnosis is proposed in this paper, building a bridge between physical features locating and machine learning methods.
    
	\item A novel unified explainable soft evaluation indicator, called XSEI, is designed to evaluate the AI-based arc fault diagnosis methods, using XAI techniques (SHAP and occlusion sensitivity). 
    
    \item We conduct an arc fault experiment using a standard arc fault test platform to demonstrate the effectiveness of XSEI, in which the arc generator follows UL1699 and GB 31143.
    \item The soft evaluation indicator is tested across several traditional machine learning methods and  deep learning methods using an arc fault dataset with different sample times and noise levels. A lightweight balance neural network is also proposed to guarantee the competitive classification accuracy and soft scores by analyzing arc faults.


\end{enumerate}

The rest of this paper is organized as follows: Section~\ref{sec:Methodology} provides the background of the utilized XAI. Section~\ref{sec:Ourmethods} explains the proposed soft evaluation indicator. Then, Section~\ref{sec:Experiments} shows all the experiments and the feature extraction attribution of all models, and Section~\ref{sec:Conclusion} concludes the paper.

\section{PRELIMINARIES}
\label{sec:Methodology}
The principle of explainable machine learning methods applied in this paper can be summarized as ``local feature attribution''. In this approach, individual predictions are explained by an attribution vector, $\psi \in \mathbb{R} ^d$. Mathematically, for any ML model $f$, input $x\in \mathbb{R}^d$, and output $y\in \mathbb{R}^c$, there is a map $g\in \mathbb{R}^{d}$ that describes the input feature importance to target class output $y_{c}$. The feature importance $g=G(x,f,y_{c})$ is calculated by a coalitional game, or heat-mapping function, based on the model being explained~\cite{chen2023algorithms}. We introduce two XAI techniques in this section: occlusion sensitivity and SHAP.

\subsection{SHapley Additive exPlanations}

The feature attributions based on SHAP depend on defining a coalitional game, or set function. The core idea of SHAP is assigning credits (contributions) to players (features) in coalition games ($G(x,f,y_{c})$) in the form of shapley values based on game theory~\cite{chen2023algorithms}. 

For a ML model $f$ and input $x\in \mathbb{R}^d$ 
, supposing that the coalition game $G$ is determined by the sets of features $s\subseteq d$ that choose to be applied in model $f$. For the i-th feature $x_{i}$, shapley values assign credit to the individual $x_{i}$ by calculating a weighted average of the $G$ increase when $x_{i}$ is in $s$ versus when $x_{i}$ does not in $s$ (a quantity known as $x_{i}$’s ‘marginal contribution’). The definition of the $x_{i}$'s shapley value $\phi_{i}$ is 
\begin{equation} \label{eq:shapley value}
\phi_{i}(G)=\sum_{s\subseteq d\setminus \left \{ x_{i} \right \} }\frac{\left | s \right |!(\left | d \right |-\left | s \right |-1 )! }{\left |d  \right | !}(G(s\cup \left \{ x_{i} \right \} )-G(s)) 
\end{equation}
It follows that when the features contribute more, they should have higher credit in the game. If the features have the same contribution, they have equal credit, and when the feature has no help, the credit is zero. 

There are several ways to remove $x_{i}$ from $d$ ($s\subseteq d\setminus \left \{ x_{i} \right \}$), including baseline sample replacement, randomly sampled replacement, and marginal distribution replacement~\cite{sundararajan2020many}. The choice of the removal approaches is determined by the complex of problems.

\subsection{Occlusion Sensitivity}
As the basic perturbation method, occlusion sensitivity analysis~\cite{zeiler2014visualizing} is normally used to generate local post-hoc explanations, specifically in the context of neural networks. By measuring the slight variation of the class score to occlusion in different regions of an input signal using small perturbations of the signal, occlusion sensitivity analysis can summarize a heat-map of the input signal based on the resultant variation of each region.

Assume that there is a mask $M$ replacing signal regions with a given baseline (blur, constant, or noise) and an ML model $f$ outputting a probability score $p\in[0,1]$. The new signal with masks can be described as $x\odot M$, where $\odot$ is the Hadamard product and the input signal $x$ is a one-dimensional data. Then, the degree of responsibility of the masked region is~\cite{valois2024occlusion}
\begin{equation} \label{eq:occlusion sensitivity analysis}
Res = 1-\frac{p(x\odot M)}{p(x)} 
\end{equation}

Assume that the masked region is defined as $R_{mask}$. When we mask all the signal, keeping nothing but $R_{mask}$ and the output prediction score $p$ is unchanged, the masked region can be considered as the core part of the signal and vice versa. Simply put,
\begin{equation} \label{eq:s}
\begin{split}
& p(x-R_{mask})=0 \Longleftrightarrow p(R_{mask})=p(x) \\
& p(x-R_{mask}')=0 \Longleftrightarrow p(R_{mask}')=0 \\
\end{split}
\end{equation}
where $R_{mask}'$ is an irrelevant area. Masking it should not affect the output score.



\section{Soft Evaluation Indicator}
\label{sec:Ourmethods}

Before diagnosing arc faults, the essential reason why ML models can identify the arc signal is that models capture the performance of the high-frequency signal in the feature pool, which will be discussed in Section~\ref{sec:Experiments}. Therefore, the ground truth features should be able to directly reflect the changes in arc current at high frequencies rather than blindly stacking useless features. Therefore, in this section, we introduce a soft evaluation indicator for both feature pool-based and deep learning-based methods, including the ground truth features definition, the soft evaluation score, and the soft evaluation process.

\subsection{Definition of Ground Truth Features}

\subsubsection{Ground truth features in feature pool}
The features currently used can be defined into three categories: original signal variants, primary, and secondary features. Original signal variants are based on signal decomposition, for example FFT, to find out the key sub-signals or corresponding properties. It can be considered as the extension of the original data instead of the statistical description. Therefore, although the data contain the most information, they are not discussed in this paper.
The primary features are the fundamental arc fault features, which are generally basic characteristics of the data, such as mean and variance. Considering that the high-frequency arc occurs under different loads and the current is periodic, the efficient features are normally based on the absolute and square value or the bias of the input signal, such as variance, entropy, range, RMS, and integral~\cite{8571267,9439848}. 
The secondary features are usually the extensions of primary features and are artificially designed, such as Skewness, Kurtosis, L1 normalization and L2 normalization. However, they lack a general specification.
Therefore, for arc fault diagnosis, five basic features (Variance, Entropy, Range, RMS, and Integral) are selected as the ground truth features set $S$. 

\subsubsection{Ground truth features in deep learning}
The ground truth features in deep learning methods are defined as the regions $\mathbf{r}=[r_1,r_2,\dots,r_N]$ where the arc fault happens. Assuming that the normal input signal at a certain type of load is $x$ and the corresponding arc fault signal is $\hat{x}$, they can be cut into $N$ independent regions respectively. The ground truth features at region $r_{n}$ can be defined as:
\begin{equation} \label{eq: Ground truth features deep learning}
r_{n} = \begin{cases}
1  & \text{ if } x_{n}\neq \hat{x}_{n}  \\
0  & \text{ if } x_{n}= \hat{x}_{n}  
\end{cases}
\end{equation}
where $n={1,2,3,\dots,N}$. $x_{n}$ and $\hat{x}_{n}$ are the n-th independent region being cut in $x$ and $\hat{x}$. 

\subsection{Definition of feature extraction score} 
For feature pool-based methods, considering the efficient features vector $\mathbf{v}$ as the ground truth and the feature extraction score of model $f_{i}$ can be defined as:
\begin{equation} \label{eq: SHAP based correct extraction}
g(f_{i}) = \frac{{S}\cap {S}_{SHAP,i}}{{S}\cup {S}_{SHAP,i}}  
\end{equation}
where ${S}_{SHAP,i}$ is the SHAP based top-5 features in model $f_{i}$ when we use Eq~\eqref{eq:shapley value}. ${S}\cap {S}_{SHAP,i}$ is the number of overlap features and ${S}\cup {S}_{SHAP,i}$ is the number of union features.
For deep learning methods, considering the arc fault ground truth features $\mathbf{r}$ in the input signals, and the feature extraction score of model $f_{i}$ can be defined as:
\begin{equation} \label{eq: occlusion based correct extraction}
g(f_{i}) = \frac{\mathbf{r}\cap \mathbf{r}_{Occlusion,i}}{\mathbf{r}\cup \mathbf{r}_{Occlusion,i}}  
\end{equation}
where $\mathbf{r}_{Occlusion,i}$ is the region marked as arc faults in model $f_{i}$ by occlusion sensitivity experiments. $\mathbf{r}\cap \mathbf{r}_{Occlusion,i}$ is the overlap of the two regions, and $\mathbf{r}\cup \mathbf{r}_{Occlusion,i}$ is the union of the two regions. The above two scores are from 0 $\sim$ 1 ($g(f_{i})\in [0,1]$). 

\subsection{Soft Evaluation Process}
Based on the definitions and the feature extraction score, soft evaluation can be used to measure the confidence of multiple AI-based arc fault diagnosis models and serve as the basis for potential ensemble learning models.

\begin{figure}[!t] 
\centering
\includegraphics[width=0.85\linewidth]{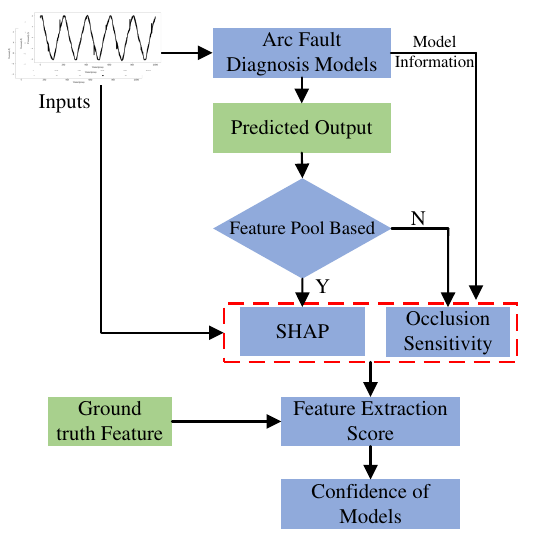}
\caption{The schematic diagram of the low-voltage soft Evaluation Process.} 
\label{fig:Soft Evaluation Process}
\end{figure}

The evaluation process is summarized in Figure~\ref{fig:Soft Evaluation Process} and described in Algorithm~\ref{alg:Soft Evaluation Process}. Depending on the model, the raw current signal is used to calculate various features or as a direct input to the model for arc fault diagnosis. The input $x$, output $\tilde{y} $, and the corresponding model $f_i$ are then used to extract the estimated key features using SHAP or occlusion sensitivity experiments. Compared with ground truth features, the feature extraction score of all models can be calculated for practitioners to determine which models are trustworthy and whether to ensemble the models.

\begin{algorithm}[!t]
{
\SetAlgoLined
\caption{$\mathrm{Soft\ Evaluation\ Process\ Algorithm}$}
\label{alg:Soft Evaluation Process}
\BlankLine
\textbf{Input}:\\
\quad $x$: input signal or features in feature pool,  \\ 
\quad $f$: AI-based arc fault diagnosis models,\\
\quad ${S}$: Ground truth features in feature pool,\\
\quad $\mathbf{r}$: Ground truth features input signals,\\
\quad $M$: The masks in occlusion sensitivity \\
\quad $L$: The number of models \\

\For{$i=1,\ldots,L$}
{
$\tilde{y} \longleftarrow f_{i}(x)$\\
\eIf {$f_{i}$ is feature pool based}
{${S}_{SHAP,i} = Top_5(SHAP(x,f_i,\tilde{y})) $\\
$g(f_{i}) = \frac{{S}\cap {S}_{SHAP,i}}{{S}\cup {S}_{SHAP,i}} $}
{ \For{$j=1,\ldots,N$}
  { $r_{Occlusion,i,j} = Occlusion Sensitivity(x\odot M_j,f_i,\tilde{y}$)}
$\mathbf{r}_{Occlusion,i} = \begin{bmatrix}
r_{Occlusion,i,1},r_{Occlusion,i,2},\dots,r_{Occlusion,i,N}
\end{bmatrix}$
$g(f_{i}) = \frac{\mathbf{r}\cap \mathbf{r}_{Occlusion,i}}{\mathbf{r}\cup \mathbf{r}_{Occlusion,i}}$
}
}
$g(f) = \begin{bmatrix}
g(f_1),g(f_2),\dots,g(f_L)
\end{bmatrix} $ \\
\textbf{Output:} \textit{Scores of all models}: $g(f)$ 
}
\end{algorithm}

\section{Experimental Results}
\label{sec:Experiments}
Before testing the explainable soft evaluation indicator, we conduct an arc fault experiment in the low-voltage electrical safety laboratory for electric appliances with/without arcing and analyze the corresponding current signals. To ensure the effectiveness of the proposed explainable soft evaluation indicator, we explore different arc fault currents under various loads, including the heating-kettle, drill, air compressor, dimmer, dehumidifier, and so on. Meanwhile, to validate the extension of XSEI in different arc fault datasets, we design two datasets. Dataset 1 is a complex dataset with more arc faults, while dataset 2 is a simple dataset with fewer arc faults.
Then, to further access the efficacy of the proposed soft evaluation indicator, we evaluate a wide range of models, including several ML models (KNN~\cite{10130787}, SVM~\cite{10141554}, CART tree~\cite{10130787}, L2/L1 norm~\cite{10499995}, SEmodel~\cite{8970332}, Lightgbm~\cite{ke2017lightgbm}, XGBoost~\cite{chen2016xgboost}, NSKSVM~\cite{11036835}, TVARF~\cite{10328664} ) and several deep learning models (AutoEncoder (AE)~\cite{9439848}, lightweight CNN (LCNN)~\cite{10018466}, ArcNet~\cite{9392282}, and ArcNN~\cite{10054597}, IFWA-1DCNN~\cite{10416657}). The ML models contain the most popular clustering, classification, and tree models, as well as ensemble models. The deep learning models are the recently proposed arc fault diagnosis models that have been proven to have good performance. We evaluate them on arc fault datasets with multi-resolution and multi-level noise. 

\subsection{Experiment Platform and Data Analysis}
\begin{figure}[!t] 
\centering
\includegraphics[width=0.9\linewidth]{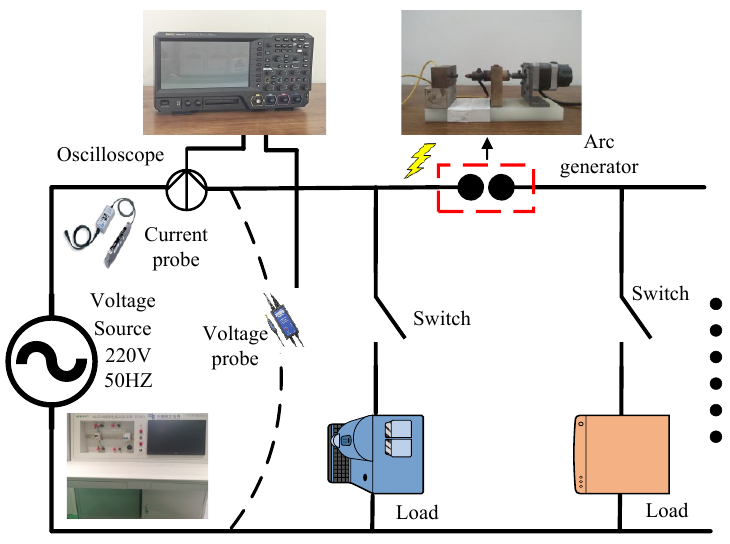}
\caption{Experimental platform of arc fault experiment.} 
\label{fig:Experimental platform}
\vspace{-0.3cm}
\end{figure}

Figure~\ref{fig:Experimental platform} shows the detailed experiment configuration, which contains an AC generator, a current probe, a voltage probe, an oscilloscope, an arc generator, and loads. The AC voltage source is operated through the CKAT10 development kit under 220V at 50Hz. The parameters analyzed within this system—specifically the currents and voltages—are captured using the Rigol Oscilloscope MSO5000 (250 Ma/S), CP8050A current probe (50A/50MHz), and Rigol voltage probe (250 MHz). The electric arc is generated by the arc generator as suggested by GB 31143 and UL 1699.

The experiments are conducted on several types of loads, including resistance, heating kettles, air compressors, vacuum cleaners, switch-mode power, dehumidifier, microwave oven, and disinfection cabinet. The arc fault dataset is obtained for a single load or combined loads. In the experiments, each fault has ten experiments, and each experiment contains 1,000,000 samples and two features (current and voltage). The arc fault occurs at a random time to simulate the real situation. The duration of each experiment is five seconds. In dataset pre-processing, we split the normal and arc fault data, cut them into several slices with a length of 10,000 (window = 10,000, step = 5,000), and downsample the data.
Due to the sampling time of $5\times 10^{-3}$ms, which means the high resolution of the data set, we performed low-resolution downsampling on the dataset such as $1\times 10^{-2}$ms, $2.5\times 10^{-2}$ms, $5\times 10^{-2}$ms, and $1\times 10^{-1}$ms for the dataset diversity. 

\begin{figure}[!t]
\centering
\subfloat[Normal current of example 1]{
\includegraphics[width=0.45\linewidth]{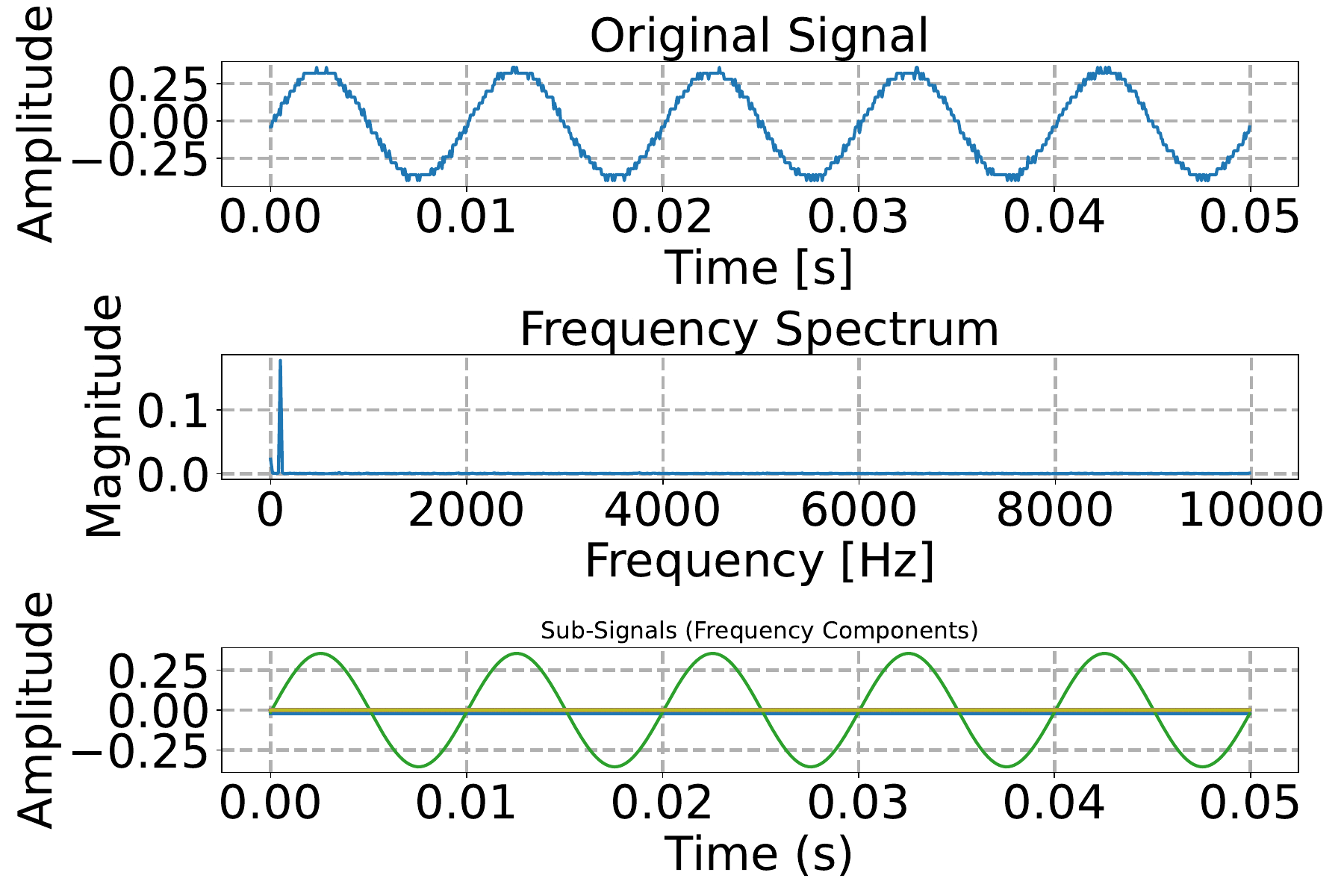}
\label{fig:Example Normal20}
} ~
\subfloat[Arc current of example 1]{
\includegraphics[width=0.45\linewidth]{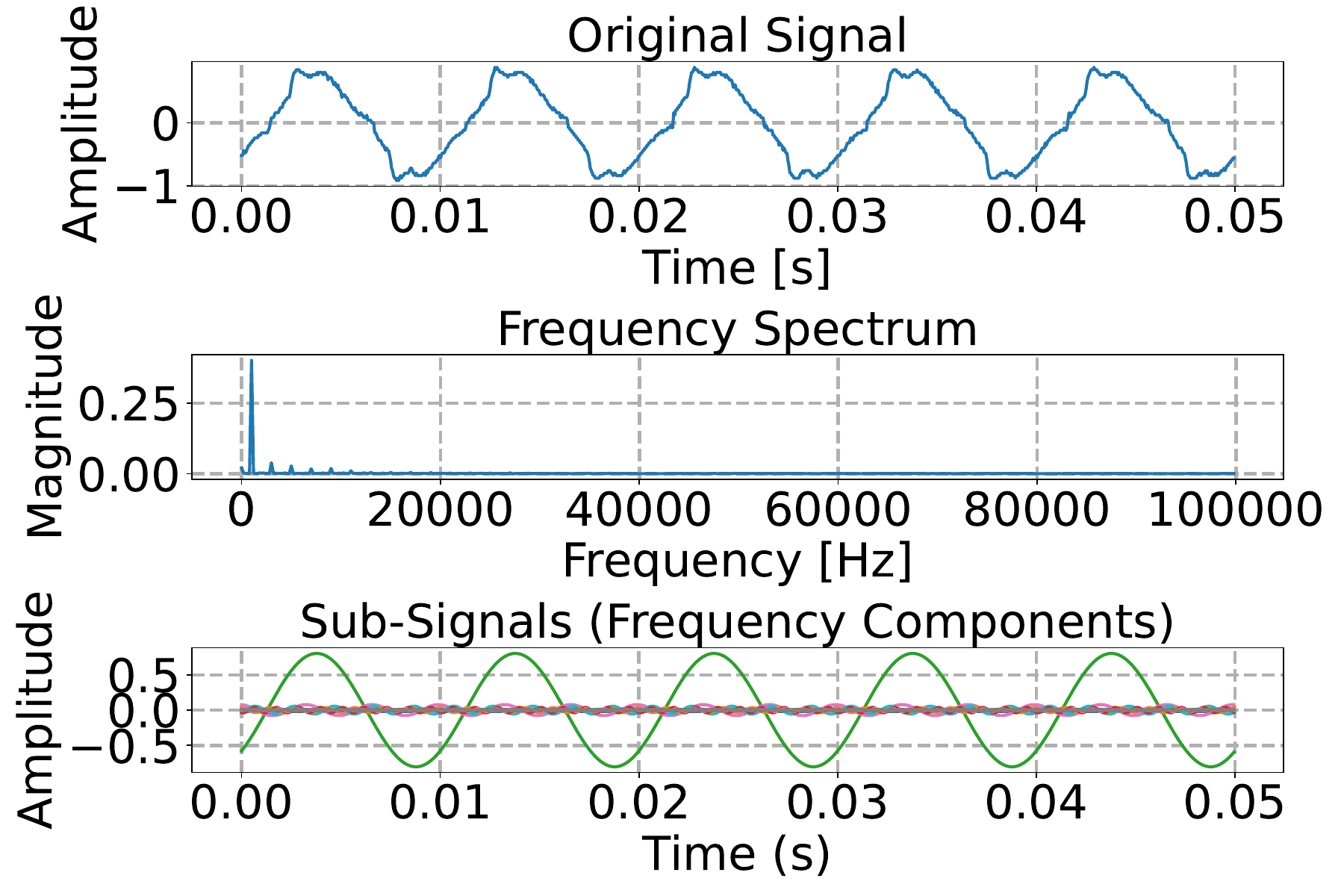}
\label{fig:Example ARC20}
} \\
\subfloat[Normal current of example 2]{
\includegraphics[width=0.45\linewidth]{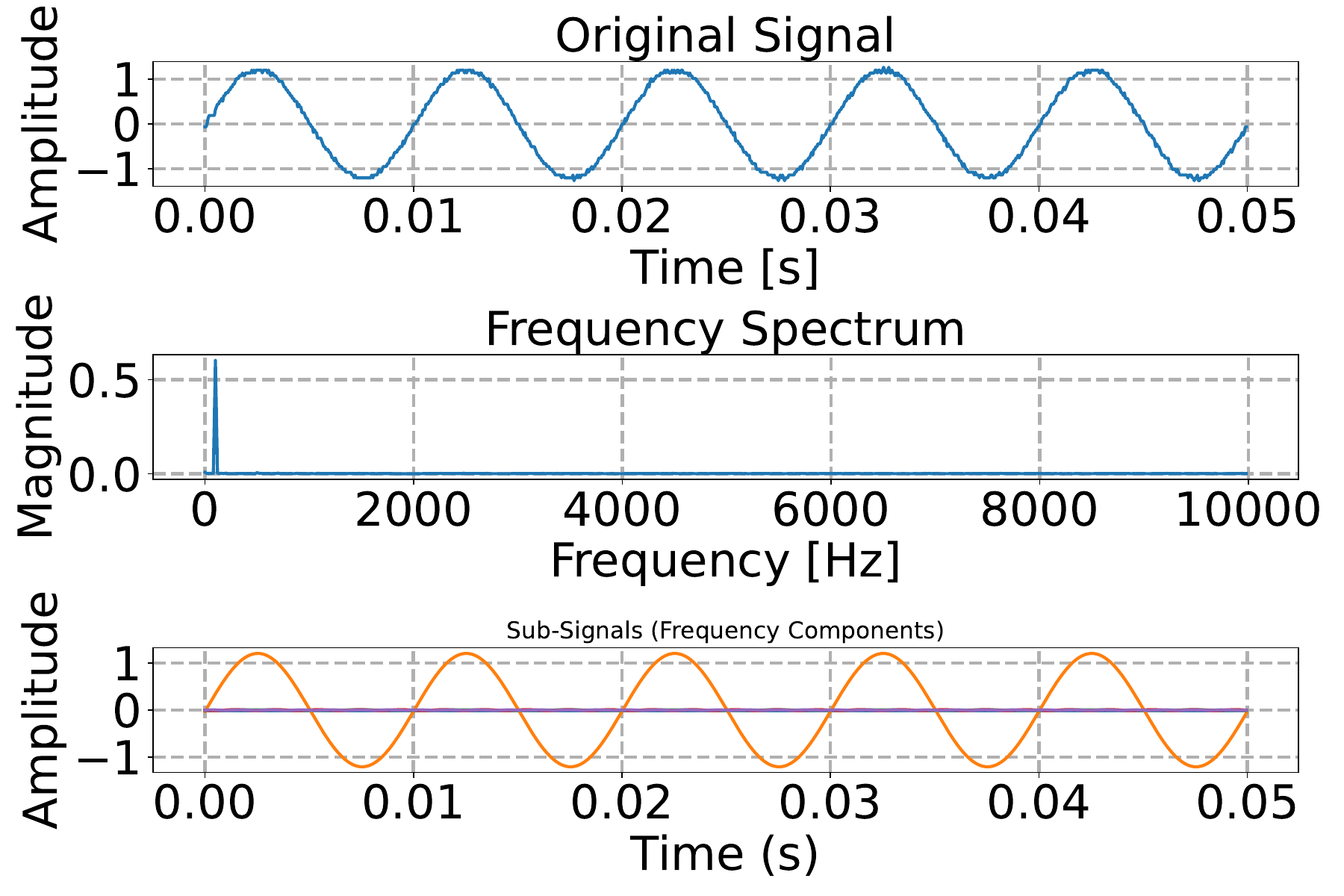}
\label{fig:Example Normal19}
} ~
\subfloat[Arc current of example 2]{
\includegraphics[width=0.45\linewidth]{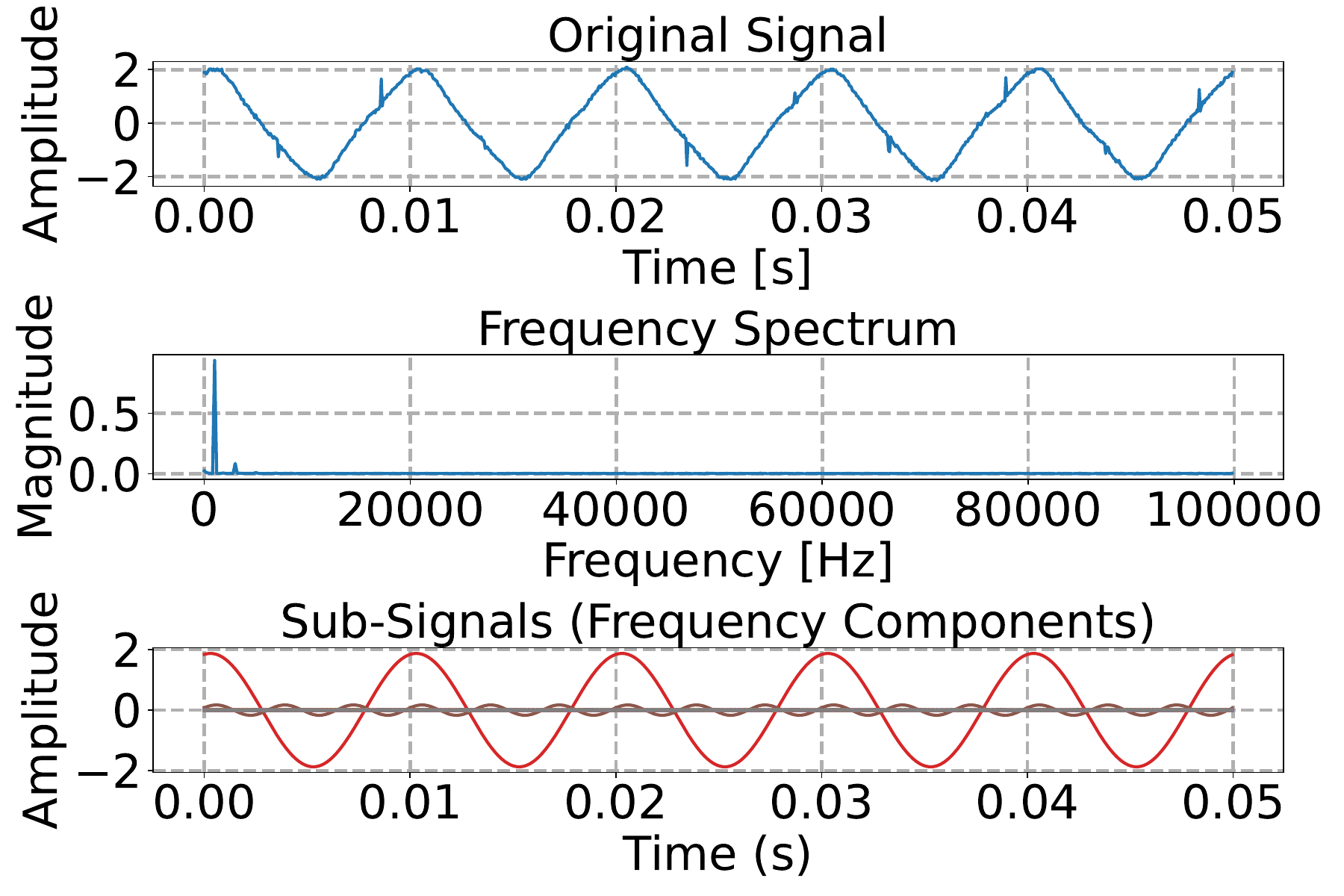}
\label{fig:Example ARC19}
} \\
\caption{Data analysis of experimental dataset. }
\label{fig:Data analysis}
\vspace{-0.3cm}
\end{figure}

Figure~\ref{fig:Data analysis} takes two examples to show the normal and arc current data (1000 samples) and performs FFT analysis. The figure contains the original signal, the detailed frequency, and the sub-signal at each frequency. The load of example 1 is a combined load, including the resistance, air compressor, and switch-mode power. For a normal current signal in Figure~\ref{fig:Example Normal20}, it has many small but high-frequency fluctuations while maintaining a sinusoidal signal due to the inherent switch-mode power. These fluctuations are hard to observe when we perform an FFT frequency spectrum unless we describe the sub-signal of each frequency. For this kind of complex load, the current changes greatly when an arc fault occurs. The current signal has completely changed into a periodic signal of other shapes instead of a sinusoidal signal. There is significant distortion in the peak value of the current, and some other high-frequency signals can be detected in both the frequency spectrum and sub-signals.
The load of example 2 is a combined load as well, including the resistance and the vacuum cleaner. Unlike Example 1, the normal current of Example 2 is more like a smooth sinusoidal signal. There are only some high-frequency noise that can be ignored. The arc fault current approximates a triangle signal with observed transient spikes, which is different from Figure~\ref{fig:Example ARC20}. The two complex current signal examples with combined loads both show differences from arc fault currents with a single load~\cite{10130787}. Whereas, the core obstacle in arc fault diagnosis is the same, that is, finding the high-frequency fault signal in the current signal without being affected by noise and sampling rate. In addition, since the laboratory experiments are relatively simple, to simulate more practical and complex engineering problems, we add additional noise with different levels of signal-to-noise ratio (SNR) to the dataset.

\subsection{Model Setup and Accuracy}
For ML models, we follow the setups of the corresponding research. For deep learning models, we train the models with a batch size of 32 using the ‘Adam’ optimizer with an initialized learning rate of $0.001$. The learning rate decays every 30 epochs, and the epoch is set to 100. The dataset is split into training, validation, and test sets with a ratio of 0.8:0.1:0.1. The experiments are implemented in 'Pytorch' using a CPU Intel i7-11800H processor at 2.3Hz and a GPU NVIDIA T600.

Table~\ref{Table: basic experiments with different models} summarizes the test accuracy of arc fault diagnosis when the sample time is $2.5\times 10^{-2}$ms and SNR is 5.
It can be seen that all models except SVM have comparative performance. When ML models are applied for fault diagnosis, test accuracy is greatly affected by model selection. The tree models, including L2/L1 norm, LightGBM, XGBoost and TVARF, all have better performance than other ML models. In contrast, all deep learning models have stable performance. The lowest test accuracy is 98.28\% (ArcNN), which is higher than the test accuracy of all ML models. However, this result-oriented analysis cannot guarantee that the model is correct and effective. Higher test accuracy cannot guarantee that the models locate the real fault features instead of finding some similar noise. 

\begin{table}[!t]
\centering\footnotesize
\caption{The basic test accuracy of arc fault diagnosis using dataset 1.}
\label{Table: basic experiments with different models}
\renewcommand{\arraystretch}{1.3}
\scalebox{0.9}{\begin{tabular}{ c| l l l l l  }
\toprule
Model & SVM & L2/L1 & KNN & SEmodel & CART  \\ 
\hline
Accuracy & 43.75\% & 92.41\% &76.11 \% &90.17\% & 90.18\%   \\
\hline
Model &  LightGBM &  XGBoost &  NSKSVM & TVARF  & AE  \\
\hline
Accuracy & 92.19\%  & 92.19\% & 83.31\% & 94.19\% & 98.65\% \\
\hline
Model & LCNN &ArcNet & ArcNN & FWA-1DCNN & - \\
\hline
Accuracy  & 99.79\% & 99.47\% & 98.28\% & 100.00\% & - \\
\bottomrule 
\end{tabular}}
\vspace{-0.3cm}
\end{table}

\subsection{Soft Evaluation} \label{evaluation}

\begin{table*}[!t]
\centering\footnotesize
\caption{The test accuracy and soft score of arc fault diagnosis using dataset 1 with different sample times.}
\label{Table: test accuracy and score with different sample times dataset 1}
\renewcommand{\arraystretch}{1.3}
\scalebox{1}{\begin{tabular}{ l| c| c| c| c| c| c| c| c| c| c| c| c }
\toprule
\multirow{2}{*}{Sample Time} & \multicolumn{2}{c|}{$5\times 10^{-3}$ms} &  \multicolumn{2}{c|}{$1\times 10^{-2}$ms} & \multicolumn{2}{c|}{$2.5\times 10^{-2}$ms} & \multicolumn{2}{c|}{$5\times 10^{-2}$ms} & \multicolumn{2}{c|}{$1\times 10^{-1}$ms} & \multicolumn{2}{c}{Average}  \\ 
\cline{2-13}
 & Acc & Score & Acc & Score & Acc & Score & Acc & Score & Acc & Score & Acc & Score \\
\hline
SVM~\cite{10141554} & 24.44\% & 0.14 & 28.89\% & 0.14 & 43.75\% & 0.14  & 47.75\% & 0.14 & 48.32\% & 0.14 & 38.36\%  & 0.14 \\
L2/L1 norm~\cite{10499995} & 91.12\% & 0.66 & 96.01\% & 1.00 & 92.42\% & 1.00 & 91.89\% & 0.45 & 86.67\% & 0.66 & 91.62\%  &  0.75 \\
KNN~\cite{10130787} & 76.67\% & 0.14 & 77.77\% & 0.33 & 76.11\% & 0.40 & 76.46\% & 0.17 & 74.67\% & 0.14 & 76.34\% & 0.24 \\
SEmodel~\cite{8970332} & 90.00\% & 1.00 & 92.22\% & 0.80 & 90.17\% & 1.00 & 91.52\% & 0.67 & 83.87\% & 0.67 & 89.56\% & 0.83 \\
CART~\cite{10130787} & 86.67\% & 0.33 & 94.44\% & 1.00 & 90.18\% & 0.67 & 88.28\% & 0.43 & 84.37\% & 0.43 & 88.79\% & 0.57 \\
LightGBM~\cite{ke2017lightgbm} & 92.22\% & 0.80 & 92.67\% & 0.67 & 92.19\% & 0.60 & 92.41\% & 0.67 & 85.99\% & 0.40 & 91.09\% & 0.63  \\
XGBoost~\cite{chen2016xgboost} & 91.11\% & 0.67 & 93.89\% & 0.43 & 92.19\% & 0.43 & 91.63\% & 0.67 & 86.66\% & 0.57 & 91.09\% & 0.55 \\
NSKSVM~\cite{11036835} & 78.89\% & 0.27 & 81.29\% & 0.20 & 85.31\% & 0.20 & 84.70\% & 0.20 & 80.97\% & 0.27 & 87.97\% & 0.23 \\
TVARF~\cite{10328664} & 92.22\% & 0.53 & 94.26\% & 0.60 & 94.19\% & 0.73 & 91.10\% & 0.60 & 87.42\% & 0.67 & 91.84\% & 0.63 \\
\hline
AE~\cite{9439848} & 88.14\% & 0.56 & 95.55\% & 0.44 & 98.65\% & 0.48 & 99.44\% & 0.48 & 99.66\% & 0.26 & 96.28\% & 0.44 \\
LCNN~\cite{10018466} & 94.44\% & 0.22 & 98.88\% & 0.26 & 99.79\% & 0.37 & 99.88\% & 0.30 & 100.00\% & 0.59 & 98.59\% & 0.34 \\
ArcNet~\cite{9392282} & 96.66\% & 0.33 & 99.07\% & 0.44 & 99.47\% & 0.44 & 99.55\% & 0.15  & 99.60\% & 0.26 & 98.78\% & 0.32 \\
ArcNN~\cite{10054597} & 83.33\% & 0.67 & 94.63\% & 0.11 & 98.28\% & 0.44 & 99.33\% & 0.59 & 99.49\% & 0.56 & 95.01\% & 0.47 \\
IFWA-1DCNN~\cite{10416657} & 95.92\% & 0.59 & 98.67\% & 0.26 & 100.00\% & 0.44  & 100.00\% & 0.33 & 99.94\% & 0.33 & 98.91\% & 0.39 \\
\bottomrule 
\end{tabular}}
\vspace{-0.3cm}
\end{table*}

\begin{table*}[!t]
\centering\footnotesize
\caption{The test accuracy and soft score of arc fault diagnosis using dataset 1 with different SNR.}
\label{Table: test accuracy and score with different SNR dataset 1}
\renewcommand{\arraystretch}{1.3}
\scalebox{0.9}{\begin{tabular}{ l| c| c| c| c| c| c| c| c| c| c| c| c| c| c }
\toprule
\multirow{2}{*}{SNR} & \multicolumn{2}{c|}{-5} &  \multicolumn{2}{c|}{-3} & \multicolumn{2}{c|}{-1} & \multicolumn{2}{c|}{1} & \multicolumn{2}{c|}{3} & \multicolumn{2}{c|}{5} & \multicolumn{2}{c}{Average}  \\ 
\cline{2-15}
 & Acc & Score & Acc & Score & Acc & Score & Acc & Score & Acc & Score & Acc & Score & Acc & Score \\
\hline
SVM~\cite{10141554} & 56.88\% & 0.14 & 57.03\% & 0.14  & 55.95\% & 0.14 & 44.27\%& 0.14 & 46.99\% & 0.14 & 47.75\%  & 0.14  & 51.47\%  & 0.14\\
L2/L1 norm~\cite{10499995} & 81.54\% & 1.00 & 81.21\% & 0.89 & 83.29\% & 0.78 & 85.12\% & 0.78 & 88.50\% & 0.89 & 91.89\%  &  0.45 & 85.25\%  & 0.80 \\
KNN~\cite{10130787} & 73.36\% & 0.30 & 72.54\% & 0.18 & 72.66\% & 0.27 & 73.74\% & 0.29 & 75.52\% & 0.30 & 76.46\% &  0.17 & 74.05\% & 0.25 \\
SEmodel~\cite{8970332} & 81.21\% & 1.00 & 82.69\% & 1.00 & 82.78\% & 0.89 & 83.82\% & 1.00 & 86.90\% & 1.00 &  91.52\% & 0.67 & 84.82\% & 0.93 \\
CART~\cite{10130787} & 76.93\% & 0.78 & 78.49\% & 0.78 & 78.50\% & 0.56 & 80.92\% & 0.59 & 85.04\% & 0.70 & 88.28\% & 0.43 & 81.36\% & 0.64 \\
LightGBM~\cite{ke2017lightgbm} & 80.21\% & 0.67 & 80.43\% & 0.78 & 83.29\% & 0.69 & 83.78\% &0.67  & 87.94\% & 0.89 & 92.41\% & 0.67 & 84.68\% & 0.73  \\
XGBoost~\cite{chen2016xgboost} & 80.35\% & 0.78 & 80.91\% & 0.67 & 83.30\% & 0.67 & 84.15\% & 0.59 & 88.20\% & 0.59 & 91.63\% & 0.67 & 84.75\% & 0.66 \\
NSKSVM~\cite{11036835} & 73.10\% & 0.20 & 75.78\% & 0.20 & 76.63\% & 0.20 & 78.09\% & 0.20 &  80.76\% & 0.20 & 84.70\% & 0.20   & 78.18\% & 0.20 \\
TVARF~\cite{10328664} & 79.53\% & 0.60 & 81.44\% & 0.53 & 83.07\% & 0.60 & 85.53\% & 0.73  & 88.06\% & 0.60 & 91.10\% & 0.60 & 84.79\% & 0.61 \\
\hline
AE~\cite{9439848} & 92.34\% & 0.33 & 96.17\% & 0.63 & 96.35\% & 0.33 & 98.14\% & 0.44 & 98.70\% & 0.41 & 99.44\% & 0.48 &  96.86\% & 0.44 \\
LCNN~\cite{10018466} & 94.43\% & 0.26 & 98.21\% & 0.30 & 99.47\% & 0.30 & 99.85\% & 0.44 & 99.88\% & 0.44 & 99.88\% & 0.30 & 99.39\% & 0.34\\
ArcNet~\cite{9392282} & 96.84\% & 0.26 & 98.77\% & 0.44 & 99.50\% & 0.30 & 99.52\% &  0.48 & 99.54\% & 0.22  & 99.55\% & 0.15 & 98.95\% & 0.31 \\
ArcNN~\cite{10054597} & 93.94\% & 0.48 & 97.32\% & 0.59 & 98.36\% & 0.52 & 98.44\% & 0.56  & 98.88\% & 0.41 & 99.33\% & 0.59 & 97.71\% & 0.53 \\
IFWA-1DCNN~\cite{10416657} & 95.61\% & 0.26 & 97.99\% & 0.48 & 99.40\% & 0.33 & 99.52\% & 0.30 & 99.89\% & 0.44 & 100.00\% &0.33  & 98.74\% & 0.36 \\
\bottomrule 
\end{tabular}}
\vspace{-0.3cm}
\end{table*}

Soft evaluation explores the soft feature extraction scores of each model at different data resolutions and noise. By choosing different sample times, the dataset can be down-sampled with various resolutions. Meanwhile, noise with different levels of signal-to-noise ratio is added to the dataset for comparison.

\subsubsection{Results using datasets with multiple sample times}
Table~\ref{Table: test accuracy and score with different sample times dataset 1} and Table~\ref{Table: test accuracy and score with different sample times dataset 2} summarize the test accuracy and soft feature extraction scores of each model on different datasets with multiple sample times and SNR=5. For ML model-based arc fault diagnosis, the large sample time limits the precision of soft feature extraction, leading to a decrease in test accuracy of all models. Due to the excellent structure of the tree models, they perform better than other models in test accuracy, especially L2/L1 norm, LightGBM, XGBoost, and TVARF. However, considering the average XSEI, the SEmodel performs better than others, instead of models with high accuracy, benefiting from the ensemble mechanism. Meanwhile, although NSKSVM improves the test accuracy of SVM through Newton-optimization, the soft feature extraction score does not improve much, only by 0.09. 
Generally, high accuracy means high interpretability, for example, the interpretability of XGBoost is always better than KNN. Whereas it is not always correct. When the accuracy is comparable, the soft feature extraction score varies depending on the choice of ML model, for example, when the sample time is $1\times 10^{-2}$, the soft scores of LightGBM and SEmodel are higher than XGBoost, while the test accuracy is lower than it. 

For deep learning models, the test accuracy is normally higher than ML models, indicating that the automatic feature extraction from raw data might be better than building an artificial feature pool. Different from ML models, as the sampling time increases (limited increase), the test accuracy of deep learning models improves, since they do not need a very small sample time to ensure the precision of detailed features. The same phenomenon is that high test accuracy does not guarantee high interpretability. The AE and ArcNN have lower average accuracy but higher scores than LCNN, ArcNet, and IFWA-1DCNN, which also occurs in each sub-experiment with different sampling times. We also notice that the test accuracy of IFWA-1DCNN quickly reaches 100\% as the sampling time becomes $2.5 \times 10^{-2}$ and $5\times 10^{-2}$ and then drops by 0.06\% slightly when the sampling time is $1 \times 10^{-1}$. This may be due to the large kernel size of the max-pooling layer, which is automatically searched by the fireworks algorithm. It prompts us to think about a problem: does a small kernel size in max-pooling conform to the characteristics of arc faults? Therefore, based on this rethinking, we proposed a lightweight balance neural network in Section~\ref{LBNN}.
Notably, the comparison between ML models and deep learning models is not fair, so this article does not mention it. 

Another interesting thing in Table~\ref{Table: test accuracy and score with different sample times dataset 1} and Table~\ref{Table: test accuracy and score with different sample times dataset 2} that the models with higher soft scores all show a rapid decay in test accuracy when the sampling time cannot meet the requirements of feature extraction such as SEmodel ($1\times 10^{-1}$), LightGBM ($1\times 10^{-1}$), AE ($5\times 10^{-3}$), and ArcNNs ($5\times 10^{-3}$). From this perspective, over-emphasizing generalization does not seem to be a good choice for task-driven AI models, as this will make the model focus on general features rather than task-based features. On the contrary, as long as the accuracy is within an acceptable range, a rapid drop in test accuracy is a sign that the model has found the key features and the model is more trustworthy.

\begin{figure}[!t]
\centering
\subfloat[SHAP value of L2/L1 norm model ($1\times 10^{-2}$ms)]{
\includegraphics[width=0.45\linewidth]{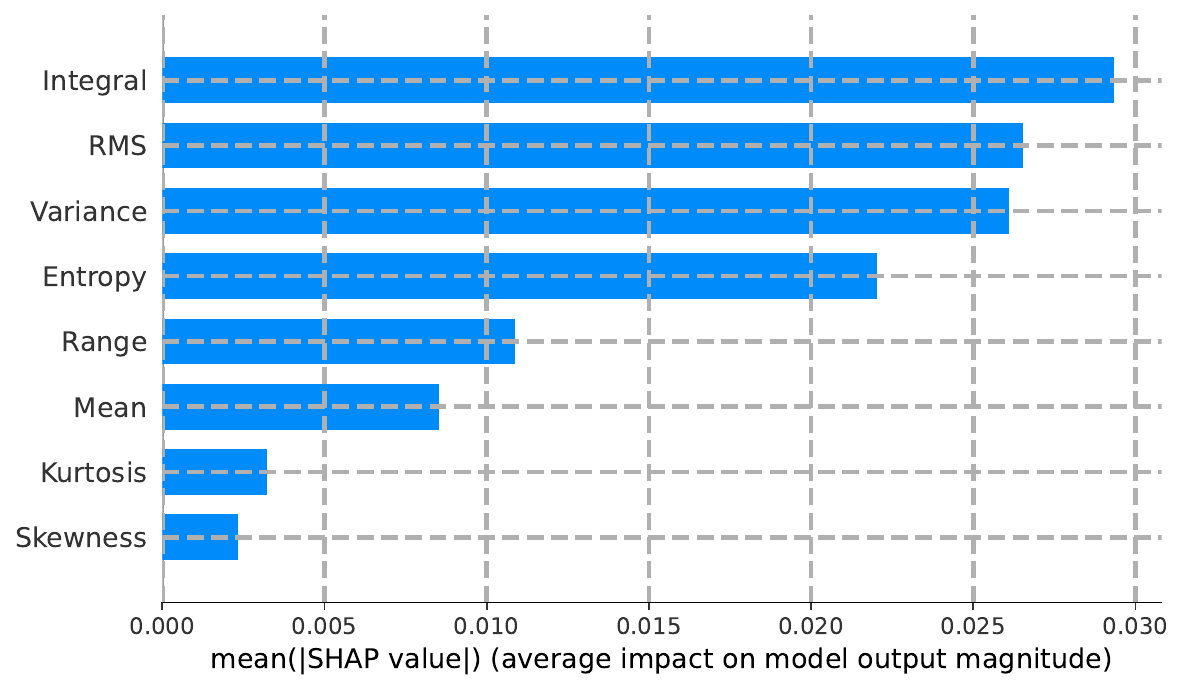}
\label{fig:SHAP_1 RF with different sample times}
} ~
\subfloat[SHAP value of XGBoost ($1\times 10^{-2}$ms)]{
\includegraphics[width=0.45\linewidth]{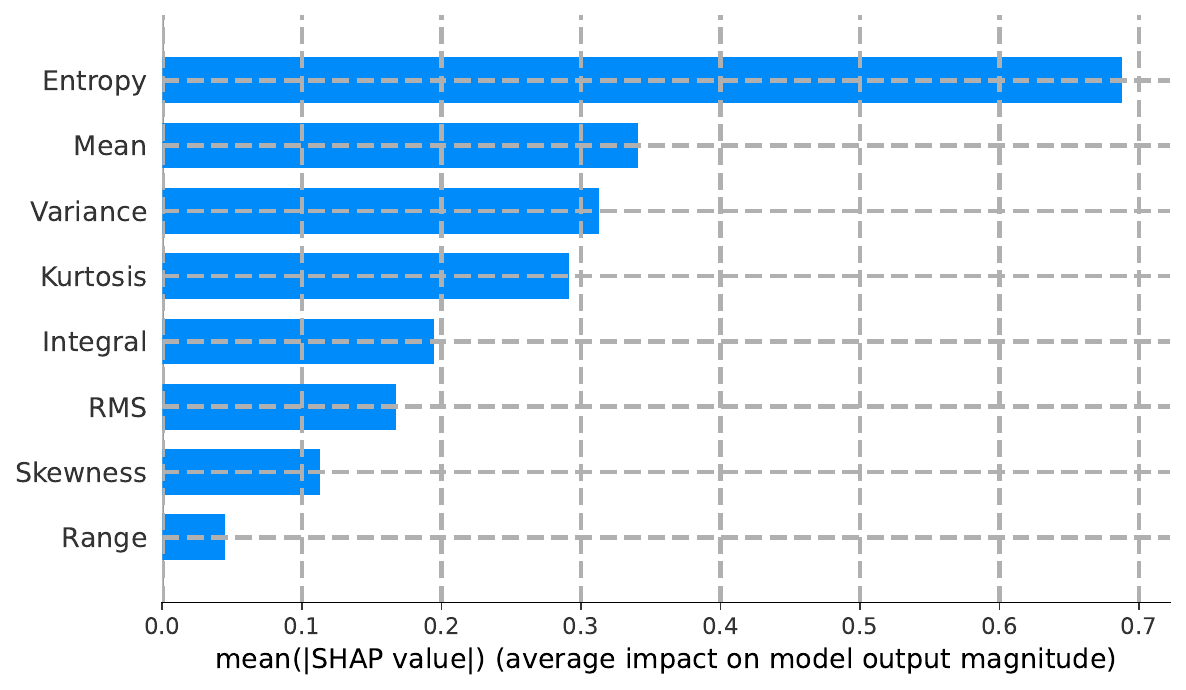}
\label{fig:SHAP_0.43 XGBoost with different sample times}
} \\
\subfloat[Occlusion experiment of ArcNN ($1\times 10^{-1}$ms)]{
\includegraphics[width=0.45\linewidth]{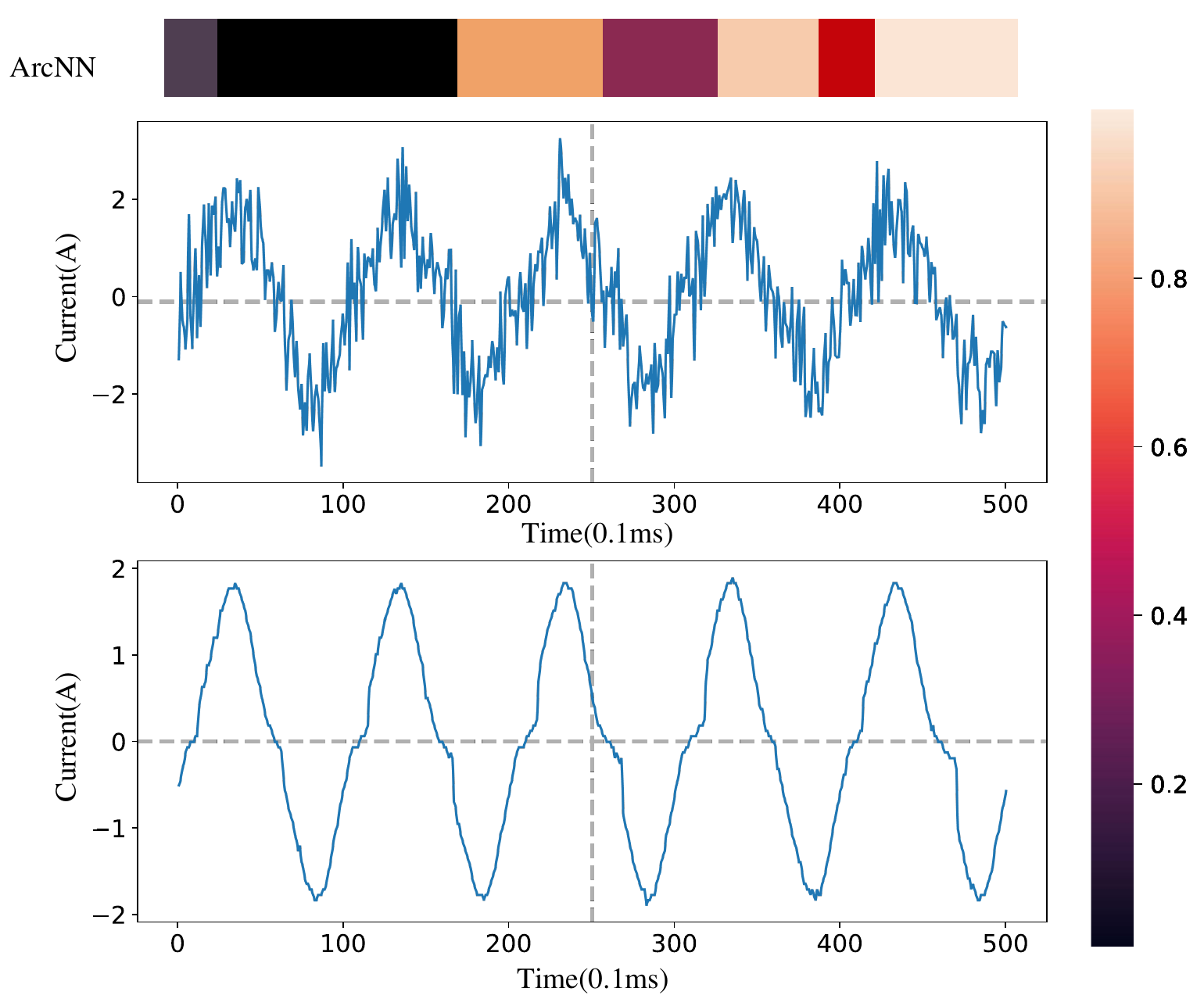}
\label{fig:Occlusion_exp for ArcNN with different sample times}
} ~
\subfloat[Occlusion experiment of ArcNet ($1\times 10^{-1}$ms)]{
\includegraphics[width=0.45\linewidth]{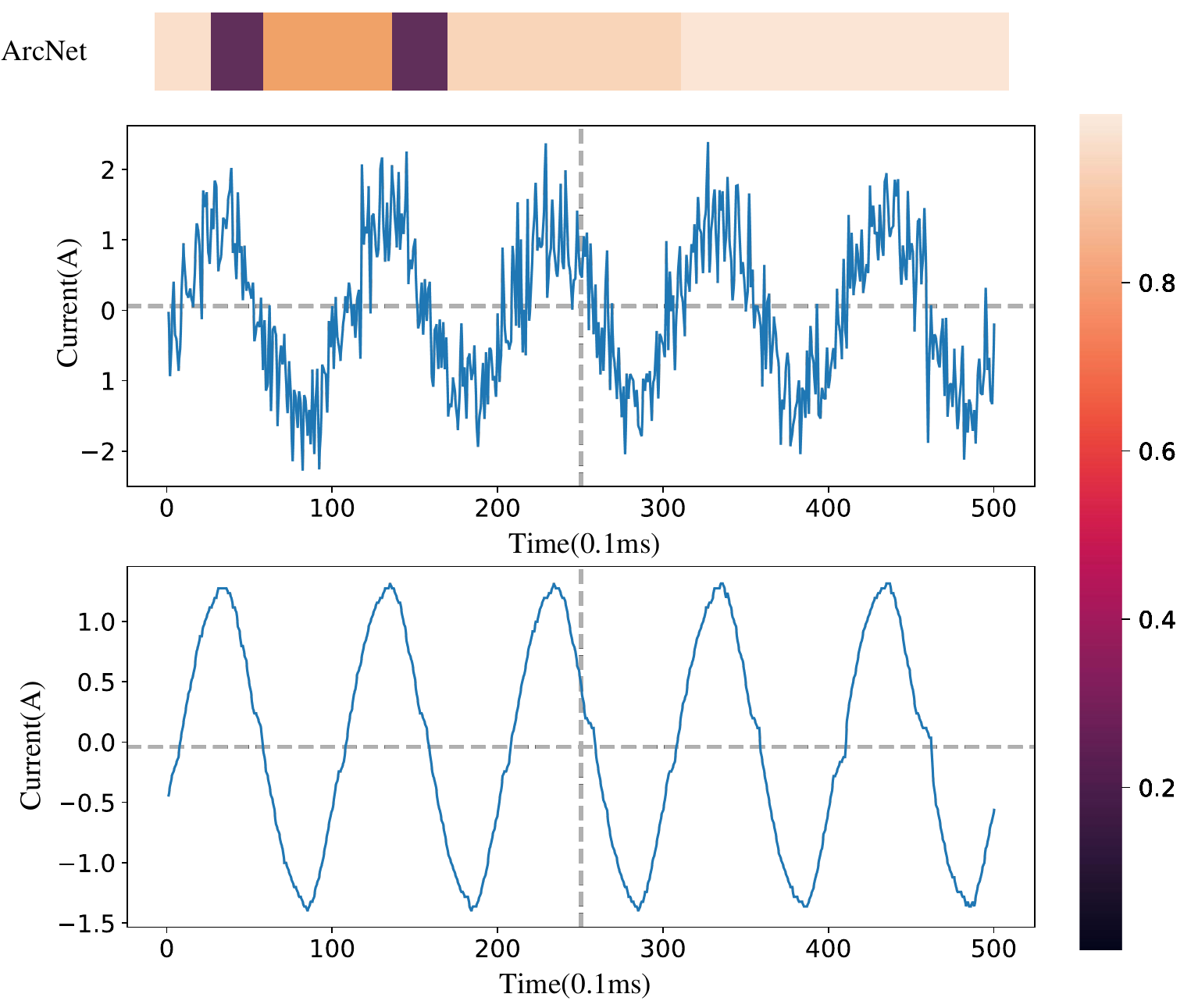}
\label{fig:Occlusion_exp for ArcNet with different sample times}
} \\
\caption{Visualization examples of soft feature extraction scores using datasets with different sample times.}
\label{fig:Visualization of soft feature extraction scores with different sample times}
\vspace{-0.3cm}
\end{figure}

Figure~\ref{fig:Visualization of soft feature extraction scores with different sample times} visualizes four soft feature extraction score experiments as examples. For the L2/L1 norm model, Integral, RMS, Variance, and Entropy have similar SHAP values, which means that the importance of the four features is similar as well. The Range is the last choice of the L2/L1 norm. The performance of XGBoost mainly relies on the Entropy instead of other features, which explains why XGBoost is less interpretable than L2/L1 norm. The SHAP values of the second and third important features are only half of the Entropy. For deep learning models, the visualization results are clearer, that is, ArcNN finds more necessary features than ArcNet. We present the original data and noisy data at the bottom and middle of the sub-figures. Although finding one arc fault feature is enough to identify the fault type, more features mean higher confidence and accuracy in other similar arc signals.

\begin{table*}[!t]
\centering\footnotesize
\caption{The test accuracy and soft score of arc fault diagnosis using dataset 2 with different sample times.}
\label{Table: test accuracy and score with different sample times dataset 2}
\renewcommand{\arraystretch}{1.3}
\scalebox{1}{\begin{tabular}{ l| c| c| c| c| c| c| c| c| c| c| c| c }
\toprule
\multirow{2}{*}{Sample Time} & \multicolumn{2}{c|}{$5\times 10^{-3}$ms} &  \multicolumn{2}{c|}{$1\times 10^{-2}$ms} & \multicolumn{2}{c|}{$2.5\times 10^{-2}$ms} & \multicolumn{2}{c|}{$5\times 10^{-2}$ms} & \multicolumn{2}{c|}{$1\times 10^{-1}$ms} & \multicolumn{2}{c}{Average}  \\ 
\cline{2-13}
 & Acc & Score & Acc & Score & Acc & Score & Acc & Score & Acc & Score & Acc & Score \\
\hline
SVM~\cite{10141554} & 66.35\% & 0.14 & 69.22\% &  0.14 & 74.20\% & 0.20 &  88.39\% & 0.20  &  88.47\% &  0.20 &  77.33\% &  0.18\\
L2/L1 norm~\cite{10499995} & 97.01\% & 0.57 & 99.51\% &  0.80 & 99.47\% &  0.80  &  99.45\% & 0.43  &  96.78\% & 0.56  &  98.44\% & 0.63 \\
KNN~\cite{10130787} &  85.12\% &  0.13 & 92.85\% & 0.33  &  92.74\% & 0.40 &  92.56\% & 0.17  &  84.44\% & 0.33  &  89.56\% & 0.27 \\
SEmodel~\cite{8970332}& 98.75\% & 0.85 & 99.38\% & 1.00 & 99.27 \% &  1.00  &  99.30\% & 0.66  &  94.47\% &  0.66 &  98.23\% & 0.83 \\
CART~\cite{10130787} & 96.31\% & 0.80 & 98.69\% & 0.46 & 98.61\% &  0.60  &  98.57\% & 0.40  &  95.01\% & 0.43  & 97.44\% &0.54  \\
LightGBM~\cite{ke2017lightgbm} & 96.45\% & 0.80 & 99.22\% & 0.58 & 99.15\% & 0.67 &  99.12\% & 0.50  &  92.31\% & 0.40  &  97.25\% & 0.59 \\
XGBoost~\cite{chen2016xgboost} & 98.01\% & 0.65 & 99.27\% & 0.43 & 99.17\% &  0.43  &  99.11\% &  0.67 &  93.10\% & 0.57  &  97.73\% & 0.55 \\
NSKSVM~\cite{11036835} & 86.49\% & 0.20 &  91.70\% & 0.20  &  92.31\% &  0.20  &  92.46\% &  0.20 &  92.39\% &  0.20 &  91.07\% & 0.20 \\
TVARF~\cite{10328664} & 97.71\% & 0.53 &  99.47\% & 0.47  &  99.36\% &  0.60  &  99.25\% & 0.47  &  97.01\% & 0.43  &  98.45\% & 0.50 \\
\hline
AE~\cite{9439848} & 96.45\% & 0.67 &  97.77\% & 0.59  &  99.49\% & 0.48   &  100.00\% & 0.52  &  100.00\% & 0.36  &  98.74\% & 0.52 \\
LCNN~\cite{10018466} & 98.60\% & 0.26 &  99.40\% & 0.30  &  99.82\% &  0.33  &  100.00\% &  0.44 &  100.00\% & 0.22  &  99.56\% &  0.21\\
ArcNet~\cite{9392282} & 98.71\% & 0.33 &  99.41\% &  0.44 &  99.74\% &  0.26  &  100.00\% & 0.59  &  100.00\% & 0.33  &  99.57\% & 0.39 \\
ArcNN~\cite{10054597} & 95.12\% & 0.59 &  97.10\% &  0.26 &  99.36\% &  0.56  &  100.00\% &  0.48 &  100.00\% &  0.67 &  98.33\% & 0.51 \\
IFWA-1DCNN~\cite{10416657} & 98.64\% & 0.44 &  99.40\% & 0.33  &  100.00\% &  0.48  &  100.00\% & 0.33  &  100.00\% & 0.27  &  99.60\% & 0.37 \\
\bottomrule 
\end{tabular}}
\vspace{-0.3cm}
\end{table*}

\begin{table*}[!t]
\centering\footnotesize
\caption{The test accuracy and soft score of arc fault diagnosis using dataset 2 with different SNR.}
\label{Table: test accuracy and score with different SNR dataset 2}
\renewcommand{\arraystretch}{1.3}
\scalebox{0.9}{\begin{tabular}{ l| c| c| c| c| c| c| c| c| c| c| c| c| c| c }
\toprule
\multirow{2}{*}{SNR} & \multicolumn{2}{c|}{-5} &  \multicolumn{2}{c|}{-3} & \multicolumn{2}{c|}{-1} & \multicolumn{2}{c|}{1} & \multicolumn{2}{c|}{3} & \multicolumn{2}{c|}{5} & \multicolumn{2}{c}{Average}  \\ 
\cline{2-15}
 & Acc & Score & Acc & Score & Acc & Score & Acc & Score & Acc & Score & Acc & Score & Acc & Score \\
\hline
SVM~\cite{10141554} & 88.62\% & 0.14 & 88.22\% &  0.20 & 87.67\% & 0.20 & 88.11\%& 0.20 & 88.24\% & 0.20 & 88.39\%  &  0.20 & 88.20\%  & 0.19 \\
L2/L1 norm~\cite{10499995} & 98.54\% & 0.80 & 98.43\% & 0.80  & 98.53\% & 0.78 & 98.98\%& 0.78 & 99.36\% & 0.89 & 99.45\%  & 0.43  & 98.88\%  & 0.75 \\
KNN~\cite{10130787} & 92.08\% & 0.20 & 91.91\% & 0.20  & 92.32\% & 0.33 & 92.45\%& 0.33 & 92.54\% & 0.29 & 92.56\%  & 0.17  & 92.31\%  & 0.25\\
SEmodel~\cite{8970332} & 97.55\% & 0.93 & 98.42\% &  1.00 & 98.50\% & 0.89 & 98.52\%& 0.88 & 99.17\% & 1.00 & 99.30\%  & 0.66  & 98.58\%  & 0.89\\
CART~\cite{10130787} & 96.06\% & 0.60 & 96.55\% & 0.66  & 97.04\% & 0.56 & 97.24\%& 0.56 & 98.68\% & 0.73 & 98.57\%  & 0.40  & 97.36\%  & 0.59 \\
LightGBM~\cite{ke2017lightgbm} & 97.32\% & 0.67 & 97.44\% &  0.78 & 97.45\% & 0.68 & 98.01\%& 0.68 & 99.04\% & 0.89 & 99.12\%  &  0.50 & 98.06\%  & 0.70 \\
XGBoost~\cite{chen2016xgboost} & 97.45\% & 0.80 & 97.53\% &  0.66 & 97.56\% & 0.68 & 98.10\%& 0.68 & 99.06\% & 0.50 & 99.11\%  & 0.67  & 98.14\%  & 0.67\\
NSKSVM~\cite{11036835} & 91.21\% & 0.14 & 91.41\% &  0.20 & 91.52\% & 0.20 & 92.03\%& 0.20 & 92.18\% & 0.20 & 92.46\%  & 0.20  & 91.80\%  & 0.19\\
TVARF~\cite{10328664} & 97.15\% & 0.67 & 97.46\% & 0.47  & 97.51\% & 0.73 & 98.07\%& 0.43 & 98.97\% & 0.73 & 99.25\%  & 0.47  & 98.07\%  & 0.58\\
\hline
AE~\cite{9439848} & 97.89\% & 0.33 & 99.01\% & 0.56  & 99.25\% & 0.63 & 99.96\%& 0.44 & 100.00\% & 0.59 & 100.00\%  & 0.52  & 99.35\%  & 0.52\\
LCNN~\cite{10018466} & 99.45\% & 0.26 & 99.93\% & 0.26  & 99.95\% & 0.37 & 99.95\%& 0.26 & 100.00\% & 0.33& 100.00\%  &  0.44 & 99.88\%  & 0.32\\
ArcNet~\cite{9392282} & 99.51\% & 0.22 & 99.89\% & 0.30  & 99.96\% &0.48 & 99.96\%& 0.22 & 100.00\% & 0.48 & 100.00\%  & 0.59  & 99.89\%  & 0.38\\
ArcNN~\cite{10054597} & 98.26\% & 0.37 & 99.15\% & 0.48  & 99.92\% & 0.52 & 99.91\%& 0.59 & 100.00\% & 0.41 & 100.00\%  & 0.48  & 99.54\%  & 0.48\\
IFWA-1DCNN~\cite{10416657} & 99.24\% & 0.16 & 99.36\% & 0.33  & 99.40\% & 0.48 & 99.94\%& 0.44 & 100.00\% & 0.44 & 100.00\%  & 0.33  & 99.67\%  & 0.36 \\
\bottomrule 
\end{tabular}}
\vspace{-0.3cm}
\end{table*}

\subsubsection{Results using datasets with multiple SNR}
Noise is unavoidable in the real-life dataset, which is different from the experimental dataset. The level of noise will directly affect the classification performance of the models. We add additional noise to our dataset, setting the SNR to -5, -3, -1, 1, 3, 5, respectively. The sample time is $5\times 10^{-2}$ms to ensure model performance distinction.

Table~\ref{Table: test accuracy and score with different SNR dataset 1} and Table~\ref{Table: test accuracy and score with different SNR dataset 2} show the test accuracy and soft score of arc fault diagnosis using a dataset with different SNR. For ML models, the performance of the tree models and the ensemble model is superior to that of other models under different SNR conditions, including NSKSVM and KNN, which is the same as in Table~\ref{Table: test accuracy and score with different sample times dataset 1} and Table~\ref{Table: test accuracy and score with different sample times dataset 2}. The sparse Kernel in NSKSVM does not make the model more interpretable, except for providing higher test accuracy.
The SEmodel still has the best average soft feature extraction score among all models, although its average test accuracy is not the highest in all experiments. Meanwhile, comprehensively considering all tables, the L2/L1 norm can have both high test accuracy and competitive soft scores, achieving a balance between performance and explainability. Although LightGBM, XGBoost and TVARF achieve acceptable performance in Table~\ref{Table: test accuracy and score with different SNR dataset 1} and Table~\ref{Table: test accuracy and score with different SNR dataset 2}, their performance on soft feature extraction scores in Table~\ref{Table: test accuracy and score with different sample times dataset 1} and Table~\ref{Table: test accuracy and score with different sample times dataset 2} is not good enough as L2/L1 norm and SEmodel. Therefore, in the arc fault diagnosis task, for experts who want to ensure the interpretability of the model, SEmodel is a better choice. However, if we pursue the tradeoff between classification accuracy and interpretability, the L2/L1 norm may be more suitable. 

For deep learning, different from test accuracy and score with different sample times, in which the larger the sampling time, the higher the accuracy, the noise in Table~\ref{Table: test accuracy and score with different SNR dataset 1} and Table~\ref{Table: test accuracy and score with different SNR dataset 2} can mask the arc fault signal, resulting in a decrease in the test accuracy. AE and ArcNN achieve better average soft feature extraction scores than LCNN, ArcNet and IFWA-1DCNN despite the lower test accuracy, which is the same as in Table~\ref{Table: test accuracy and score with different sample times dataset 1} and Table~\ref{Table: test accuracy and score with different sample times dataset 2}. Comprehensively considering test accuracy and score with different sample times and different SNRs, IFWA-1DCNN seems to strike a balance between test accuracy and scores with moderate performance, which shows the potential of large kernel sizes in the max-pooling layers. In general, deep learning is a better choice for arc fault diagnosis when considering classification accuracy, model interpretability, and visualization of fault location.

Another reason to choose deep learning over ML for arc fault diagnosis is that the fast drop in test accuracy is not obvious or too early in ML experiments, including SEmodel and L2/L1 norm models. There is a 3\% drop in test accuracy when the SNR is changed from 3 to 1. For a model that accurately locates a real arc fault, the test accuracy should drop significantly when the noise is sufficient to interfere with the model. Meanwhile, we hope that this phenomenon will occur as late as possible.
Contrary to ML models, we notice that there is a 4\% drop in test accuracy when we use AE and ArcNN in SRN=-5 and SNR=-3, which is totally different from the performance of LCNN and ArcNet. During the process, the test accuracy drop of IFWA-1DCNN is not obvious enough. This phenomenon can explain why AE and ArcNN can achieve high soft feature extraction scores as well. 

\begin{figure}[!t]
\centering
\subfloat[SHAP value of SEmodel (SNR=-3)]{
\includegraphics[width=0.45\linewidth]{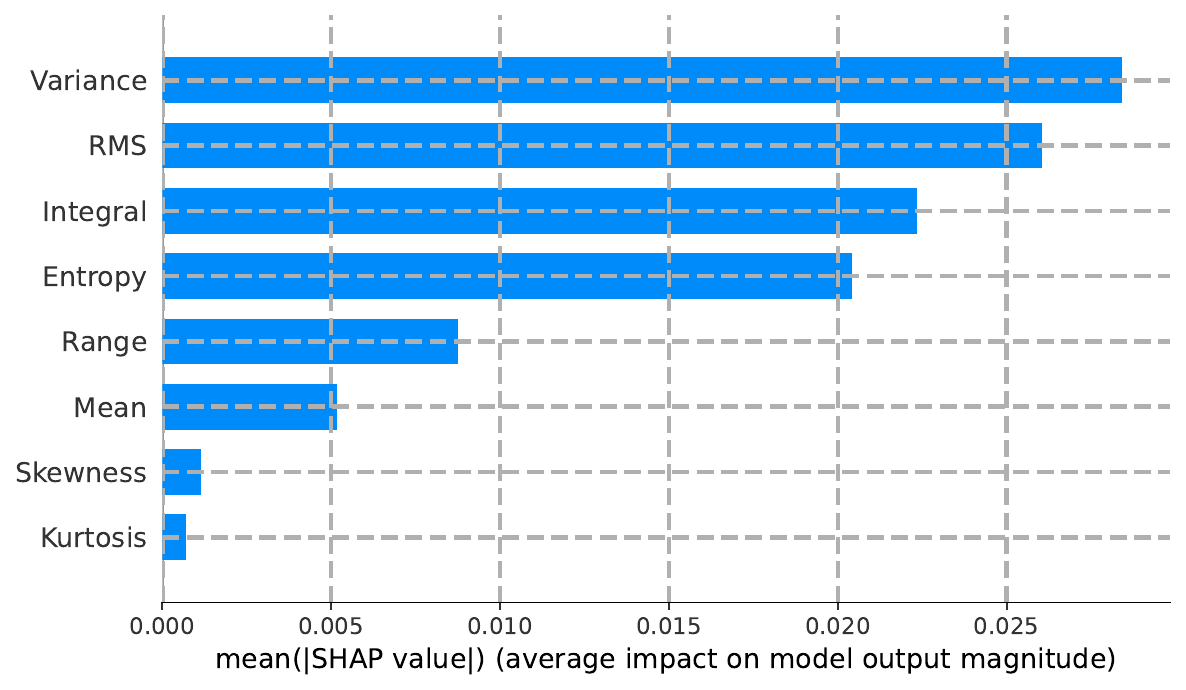}
\label{fig:shap_summary_ET_SNR with different SNR}
} ~
\subfloat[SHAP value of L2/L1 norm (SNR=-3)]{
\includegraphics[width=0.45\linewidth]{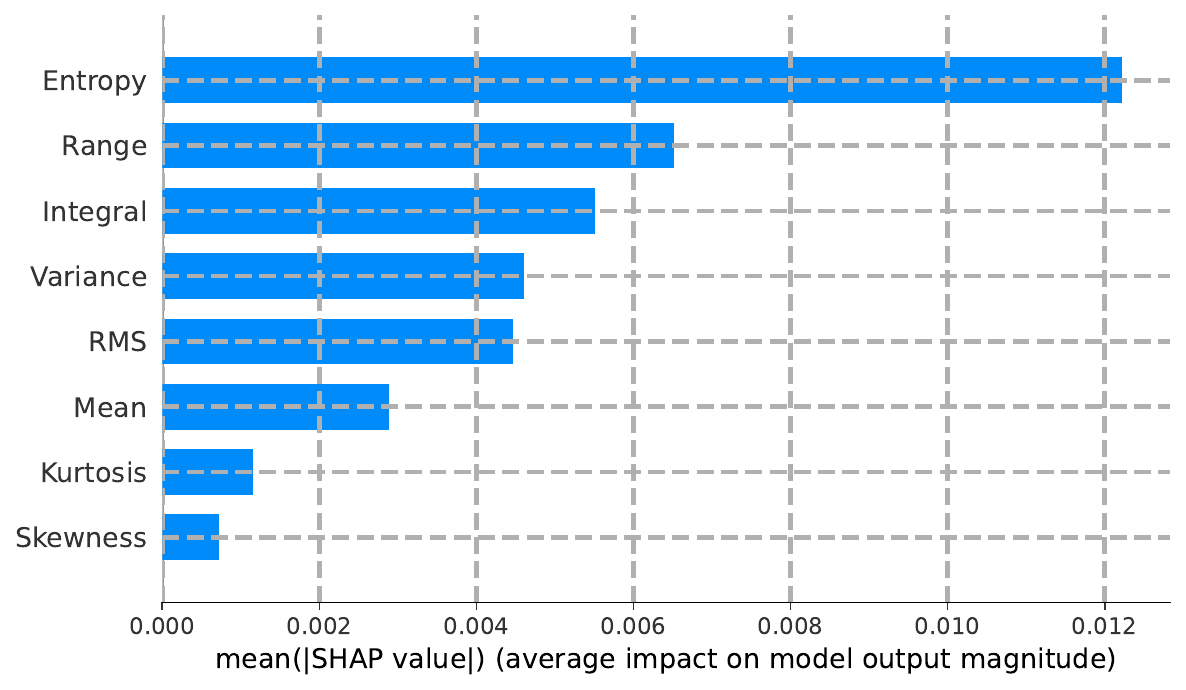}
\label{fig:shap_summary_RF_SNR with different SNR}
} \\
\subfloat[Occlusion experiment of AE (SNR=-3)]{
\includegraphics[width=0.45\linewidth]{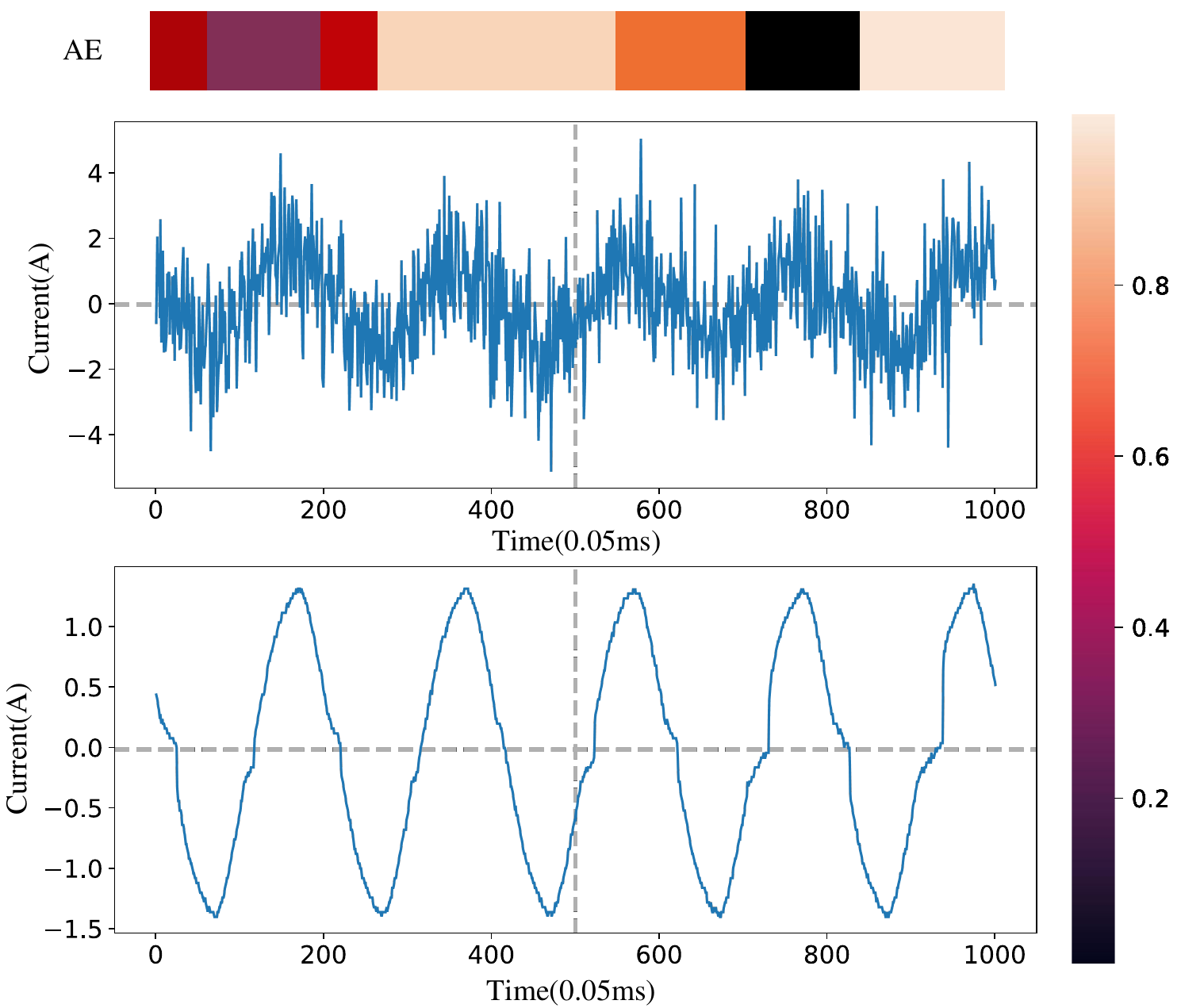}
\label{fig:Occlusion_exp for AE with different SNR}
} ~
\subfloat[Occlusion experiment of LCNN (SNR=-3)]{
\includegraphics[width=0.45\linewidth]{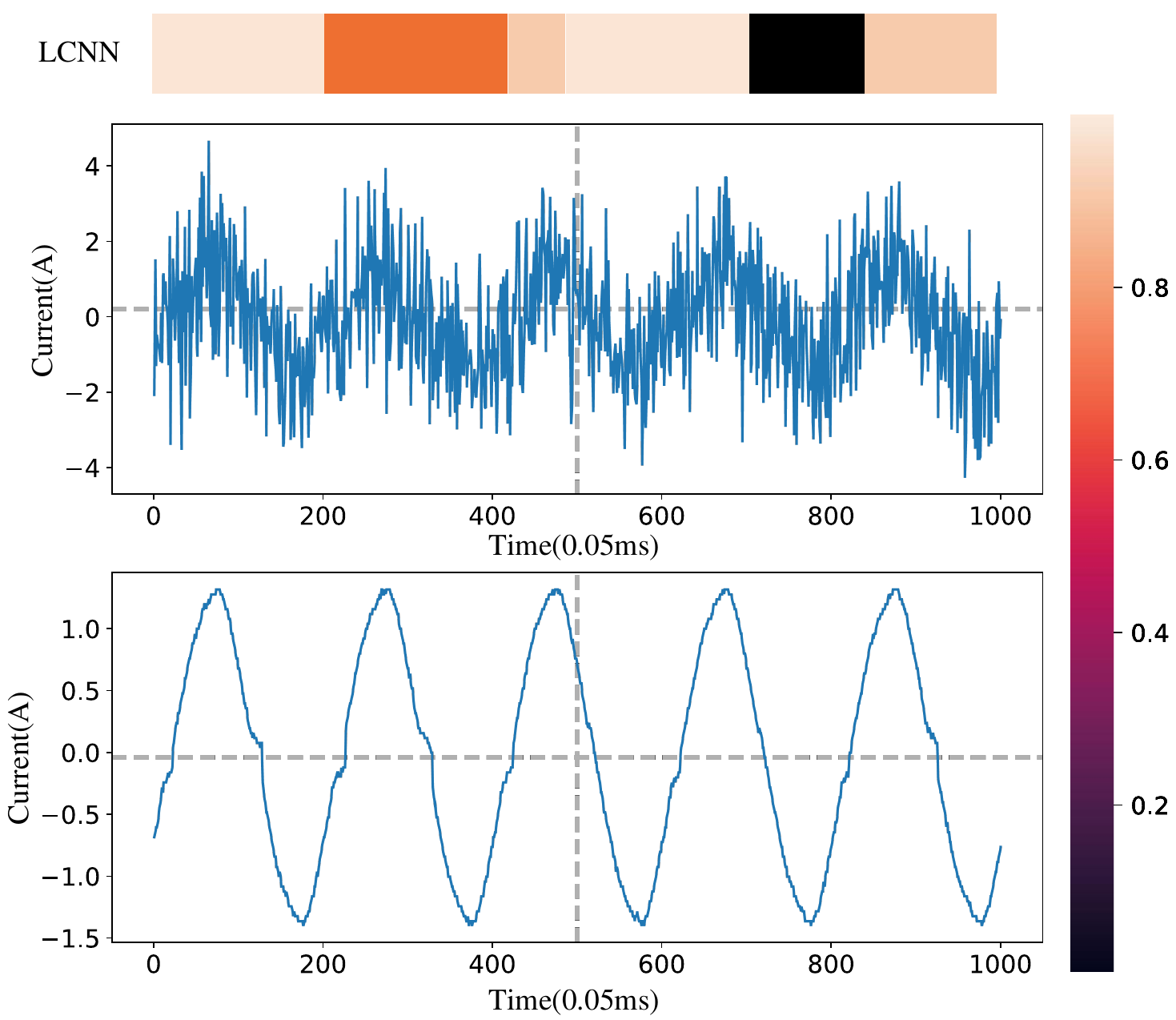}
\label{fig:Occlusion_exp for LCNN with different SNR}
} \\
\caption{Visualization examples of soft feature extraction scores using datasets with different SNR.}
\label{fig:Visualization of soft feature extraction scores with different SNR}
\vspace{-0.3cm}
\end{figure}

Figure~\ref{fig:Visualization of soft feature extraction scores with different SNR} shows four different soft feature extraction score experiments as examples. The shape values of SEmodel and L2/L1 norm explain why SEmodel performs better than L2/L1 norm in soft feature extraction scores. As depicted in Figure~\ref{fig:shap_summary_ET_SNR with different SNR} and Figure~\ref{fig:shap_summary_RF_SNR with different SNR}, SEmodel relies on four main features to classify the arc faults, while L2/L1 norm focuses more on `Entropy'. Although other main features are considered as the top-5 features as well in Figure~\ref{fig:shap_summary_RF_SNR with different SNR}, their importance is far less than `Entropy', which is opposite to SEmodel in Figure~\ref{fig:shap_summary_ET_SNR with different SNR}. The occlusion experiments verify that AE can locate the arc faults better than LCNN, especially in the presence of obvious noise. The two models both find two arc faults as the main features for fault diagnosis, which is colored as 'black'. Whereas, the difference in the interpretability of the two models comes from the identification of other fault signals.
In Figure~\ref{fig:Occlusion_exp for AE with different SNR}, near half arc faults in the signal are detected and identified as important components for arc fault diagnosis, which is different from LCNN in Figure~\ref{fig:Occlusion_exp for LCNN with different SNR}. The darker the color, the greater the impact of the fault on arc fault diagnosis.

\subsection{Lightweight Balance Neural Network} \label{LBNN}

\begin{table}[!t]
\centering\footnotesize
\caption{The PRM of deep learning methods. \#PRM denotes the number of parameters.}
\label{Table: The PRM of deep learning methods}
\renewcommand{\arraystretch}{1.3}
\scalebox{1}{\begin{tabular}{ l| c c c}
\toprule
Models & AE~\cite{9439848} & LCNN~\cite{10018466} & ArcNet~\cite{9392282}   \\
\hline
\#PRM & 32.75M & 3.23M & 4.23M  \\
\hline
Models & ArcNN~\cite{10054597} & IFWA-1DCNN~\cite{10416657} & LBNN  \\
\hline
\#PRM  & 0.14M & 1.94M  & 1.02M  \\
\bottomrule 
\end{tabular}}
\end{table}

\begin{table}[!t]
 \centering
 \caption{Lightweight Balance Neural Network}
 \label{tab:Lightweight Balance Neural Network}
 \renewcommand{\arraystretch}{1.3}
 \begin{tabular}{c|c|c}
 \toprule
  Layer & Parameters & Activation \\
 \hline
 Convolution ($5\times1$) & Filter size = 6, Padding = 2  & ReLU \\
 \hline
 AvgPooling & Kernel size = 2, Stride = 2  &-- \\ 
 \hline
  Convolution ($3\times1$) & Filter size = 16, Padding = 2  & ReLU \\
 \hline
 AvgPooling & Kernel size = 2, Stride = 2  &-- \\ 
 \hline 
 \multicolumn{3}{c}{Flatten layer} \\
 \hline
 Fully connected & Size = 256 & ReLU \\
 \hline
 Fully connected & Size = 16 & Softmax \\
 \bottomrule 
 \end{tabular}
 \vspace{-0.3cm}
 \end{table}

\begin{table*}[h]
\centering\footnotesize
\caption{The test accuracy and soft score of ArcNet and IFWA-1DCNN using average-pooling.}
\label{Table: The test accuracy and soft score of ArcNet and IFWA-1DCNN using average-pooling}
\renewcommand{\arraystretch}{1.3}
\scalebox{1}{\begin{tabular}{ l| c| c| c| c| c| c| c| c| c| c| c| c }
\toprule
\multirow{2}{*}{Sample Time} & \multicolumn{2}{c|}{$5\times 10^{-3}$ms} &  \multicolumn{2}{c|}{$1\times 10^{-2}$ms} & \multicolumn{2}{c|}{$2.5\times 10^{-2}$ms} & \multicolumn{2}{c|}{$5\times 10^{-2}$ms} & \multicolumn{2}{c|}{$1\times 10^{-1}$ms} & \multicolumn{2}{c}{Average}  \\ 
\cline{2-13}
 & Acc & Score & Acc & Score & Acc & Score & Acc & Score & Acc & Score & Acc & Score \\
\hline
ArcNet~\cite{9392282} & 94.44\% & 0.44 & 100.00\% & 0.48 & 99.93\% & 0.41  &  99.96\% &  0.26 &  99.96\% & 0.33 &  98.86\% &  0.38 \\
IFWA-1DCNN~\cite{10416657} & 94.07\% & 0.63 & 99.26\% & 0.52 & 100.00\% & 0.56  &  100.00\% & 0.33  &  99.98\% &  0.44&  98.66\% &  0.50 \\
\hline
\multirow{2}{*}{SNR} & \multicolumn{2}{c|}{-5} &  \multicolumn{2}{c|}{-3} & \multicolumn{2}{c|}{-1} & \multicolumn{2}{c|}{1} & \multicolumn{2}{c|}{3} & \multicolumn{2}{c}{Average}  \\ 
\cline{2-13}
 & Acc & Score & Acc & Score & Acc & Score & Acc & Score & Acc & Score & Acc & Score \\
\hline
ArcNet~\cite{9392282} & 98.88\% & 0.22 & 99.14\% & 0.48 & 99.81\% & 0.46  &  99.88\% & 0.33  &  100.00\% & 0.19 &  99.72\% & 0.34  \\
IFWA-1DCNN~\cite{10416657} & 98.99\% & 0.44 & 99.42\% & 0.44 & 99.51\% & 0.37  & 99.70 \% &  0.41 & 100.00\% & 0.56 &  99.52\% &  0.44 \\

\bottomrule 
\end{tabular}}
\vspace{-0.3cm}
\end{table*}

\begin{table}[!t]
\centering\footnotesize
\caption{The test accuracy and soft scores of LBNN}
\label{Table: test accuracy and soft scores of LBNN}
\renewcommand{\arraystretch}{1.3}
\scalebox{1}{\begin{tabular}{ l| c| c |c| c| c  }
\toprule
\multirow{2}{*}{Sample time} & \multicolumn{2}{c|}{Indicator} & \multirow{2}{*}{SNR} & \multicolumn{2}{c}{Indicator} \\ 
\cline{2-3} \cline{5-6}
  &  Acc & Score &  &  Acc & Score  \\
\hline
 $5\times 10^{-3}$ms & 85.56\% & 0.44  &  -5 & 98.03\% & 0.30  \\
\hline
 $1\times 10^{-2}$ms &  93.33\%  & 0.56 & -3 & 99.22\% & 0.56 \\
\hline
 $2.5\times 10^{-2}$ms & 95.36\% & 0.36 & -1 & 99.40\% & 0.41 \\
 \hline
 $5\times 10^{-2}$ms & 99.22\% & 0.56  & 1 & 99.67\% & 0.41\\
 \hline
 $1\times 10^{-1}$ms & 99.28\% & 0.39 & 3 & 99.93\% & 0.48 \\
 \hline
 Average & 94.55\% & 0.46 & Average & 99.25\% & 0.43 \\
\bottomrule 
\end{tabular}}
\vspace{-0.3cm}
\end{table}

Inspired by the good performance of IFWA-1DCNN in balancing the average test accuracy and the soft feature extraction score, we propose another lightweight balance neural network for arc fault diagnosis. We notice that the arc fault features are usually a small area of distortion rather than an instantaneous mutation, which means that the small kernel size in max-pooling may be contrary to the arc fault characteristics. Meanwhile, the larger kernel size of the max-pooling layer in IFWA-1DCNN can help models improve the test accuracy, which confirms the above idea. We also notice that the model does not need to be complex to achieve good results, for example, IFWA-1DCNN. Therefore, average pooling should be more practical than max pooling or multiple convolution layers. Table~\ref{tab:Lightweight Balance Neural Network} shows the utilized architecture of the lightweight balance neural network. Table~\ref{Table: The PRM of deep learning methods} shows the parameters of all deep learning-based arc fault diagnosis models. Except for AE, all other models are lightweight, especially ArcNN in which dilated convolution layers are used.

We test the LBNN using a dataset with different sample times with SNR = -3 and multiple SNRs with sample time = $5\times 10^{-2}$ms as well to demonstrate the effectiveness of the model. To ensure model performance distinction, the SNR = -3 instead of 5 when the sample time is different. As can be seen in Table~\ref{Table: test accuracy and soft scores of LBNN}, the LBNN can guarantee the comparative test accuracy and soft scores among all experiments. The fast drop,which is observed in both tables about the test accuracy and score with different sample times and different SNRs, appears as well in Table~\ref{Table: test accuracy and soft scores of LBNN}. When we use different SNRs, the drop happens when SNR=-9, which is not included in the table.

\subsection{The Influence of Average-pooling}

According to the discovery in Section~\ref{evaluation} and Section~\ref{LBNN}, to explore the influence of the average-pooling layers and avoid the huge change in model size, we modified two deep learning models (ArcNet and IFWA-1DCNN) using average-pooling to replace max-pooling, since there are pooling layers in the original structures.

The comparisons can be divided into two aspects. First, for data with little noise and different sample times, the average-pooling layer helps the models find the real arc faults at the expense of a slight decrease in average accuracy. Specifically, this decrease is caused by the rapid drop in test accuracy at sampling time $5\times 10^{-3}ms$. This phenomenon remains consistent across other models with high soft feature extraction scores, indicating the effectiveness of our proposed XSEI. Second, for noisy data with certain sample times, the average-pooling layer helps the models improve the test accuracy and the soft feature extraction scores, locating the real faults. The improvement of test accuracy benefits from the shift of the rapid drop phenomenon. It changes from SNR=-5 and a 2.5\% accuracy drop to SNR=-9 and a 4\% accuracy drop, which is not included in Table~\ref{Table: The test accuracy and soft score of ArcNet and IFWA-1DCNN using average-pooling}. However, the improvement in soft feature extraction scores is sufficient to reflect the model's feature extraction ability.

\section{Conclusion}\label{sec:Conclusion}

ML and deep learning models have achieved outstanding performance in arc fault diagnosis. Whereas, the inherent problem is how to determine whether these models really focus on the real arc fault. In this light, this paper summarizes the novel ML and deep learning fault diagnosis models in recent years and proposes an explainable soft evaluation indicator to evaluate the AI-based models. We also proposed a lightweight balanced neural network to guarantee competitive classification accuracy and soft score. To demonstrate the effectiveness of the soft evaluation indicator, we performed an arc fault experiment using a standard arc fault test platform. The experiments across seven traditional machine learning methods and four deep learning methods using an arc fault dataset with different sample times and noise levels are conducted as well. We show that (i) high test accuracy does not guarantee high model interpretability. (ii) An easy way to find a model with good interpretability is to find the fast drop in accuracy when using data of different precisions. (iii) For arc fault diagnosis, average pooling layer is a more efficient choice than max pooling layer. We need to highlight that the proposed XSEI is not in conflict with accuracy. Instead, it can be regarded as a supplement to the single accuracy-based arc fault evaluation system, especially when the model achieves acceptable classification accuracy. 
The main limitation of this method is the multiple combinations of classifiers and XAI techniques, which require a lot of expertise.


\end{document}